\documentclass[10pt,letterpaper]{article}
\usepackage[top=0.85in,left=2.75in,footskip=0.75in,marginparwidth=2in]{geometry}

\usepackage[utf8]{inputenc}

\usepackage{cite}

\usepackage{nameref,hyperref}

\usepackage[right]{lineno}

\usepackage{microtype}
\DisableLigatures[f]{encoding = *, family = * }

\raggedright
\usepackage{changepage}

\usepackage[aboveskip=1pt,labelfont=bf,labelsep=period,singlelinecheck=off]{caption}

\makeatletter
\renewcommand{\@biblabel}[1]{\quad#1.}
\makeatother

\usepackage{lastpage,fancyhdr,graphicx}
\usepackage{epstopdf}
\fancyheadoffset[L]{2.25in}
\fancyfootoffset[L]{2.25in}

\usepackage{color}

\definecolor{Gray}{gray}{.25}

\usepackage{graphicx}
\usepackage{algorithmic}
\usepackage{algorithm}
\usepackage{multirow}
\usepackage{amsmath}
\usepackage{amssymb}
\newtheorem{theorem}{Theorem}[section]

\usepackage{subfig}

\usepackage{sidecap}

\usepackage{wrapfig}
\usepackage[pscoord]{eso-pic}
\usepackage[fulladjust]{marginnote}
\reversemarginpar

\DeclareUnicodeCharacter{00A0}{ }

\begin{document}
\vspace*{0.35in}

% sharp symbol
%newcommand{CS}{Cnolinebreak[4]hspace{-0.05em}raisebox{.4ex}{relsize{-2}{textbf{#}}}}

% title goes here:
\begin{flushleft}
{\Large
\textbf\newline{Adaptive Online Sequential ELM for Concept Drift Tackling }
}
\newline
% authors go here:
\\
Arif Budiman,
Mohamad Ivan Fanany,
Chan Basaruddin,
%Author 4\textsuperscript{1},
%Author 5\textsuperscript{2},
%Author 6\textsuperscript{2},
%Author 7\textsuperscript{1,*}
\\
\bigskip
\bf{} Machine Learning and Computer Vision Laboratory \\Faculty of Computer Science, Universitas Indonesia\\

\bigskip
* arif.budiman21@ui.ac.id

\end{flushleft}

\providecommand{\keywords}[1]{\textbf{\textit{Keywords---}} #1}

\section*{Abstract}
A machine learning method needs to adapt to over time changes in the environment. Such changes are known as concept drift. One approach to concept drift handling is by feeding the whole training data set once again into a learning machine for retraining. Another approach is by rebuilding an ensemble classifiers to adapt to a new training data set. In either approach, retraining or rebuilding classifiers are expensive and not practical. In this paper, we propose an enhancement of Online-Sequential Extreme Learning Machine (OS-ELM) and its variant Constructive Enhancement OS-ELM (CEOS-ELM) by adding an adaptive capability for classification and regression problem. The scheme is named as Adaptive OS-ELM (AOS-ELM). It is a single classifier scheme that works well to handle real drift, virtual drift, and both drifts occurred at the same time (hybrid drift). The AOS-ELM also works well for sudden drift as well as recurrent context change type. The scheme is a simple unified method implemented in simple lines of code. We evaluated AOS-ELM on regression and classification problem by using various public dataset widely used for concept drift verification from SEA and STAGGER; and other public datasets such as MNIST and USPS. 
Experiments show that our method gives higher kappa value compared to the multi-classifier ELM ensemble. Even though AOS-ELM in practice does not need hidden nodes increase, we address some issues related to the increasing of the hidden nodes such as error condition and rank values. We propose to take the rank of the pseudo inverse matrix as an indicator parameter to detect 'under-fitting' condition. 

\bigskip

\noindent\keywords{adaptive, concept drift, extreme learning machine, online sequential.}

% now start line numbers
%\linenumbers

% the * after section prevents numbering
\section*{Introduction}
\label{Introduction}
Data stream mining is a data mining technique, in which the trained model is updated whenever new data arrive. However, the trained model must work in dynamic environments, where a vast amount of data is not only continuously generated but also keep changing. This challenging issue  is known as concept drift \cite{Gama:2014:SCD:2597757.2523813}, in which the statistical properties of the input attributes and target classes shifted over time. Such shifts can make the trained model becoming less accurate. These methods pursue an accurate, simple, fast and flexible way to retain classification performance when the drift occurs. Ensemble classifier is a well-known way to retain the classification performance. The combined decision of many single classifiers (mainly using ensemble members diversification) is more accurate than single classifier \cite{6779381}. However, it has higher complexity when handling multiple (consecutive) concept drifts. 

One of the popular machine learning methods is Extreme Learning Machine (ELM) introduced by Huang,  \textit{et. al.} \cite{huang2006extreme}\cite{journals/tsmc/HuangZDZ12}\cite{s12559-014-9255-2}, \cite{Huang201532}
\cite{ELM_filling_gap}. The ELM is a Single Layer Feedforward Neural Network (SLFN) with fast learning speed and good generalization capability. 

In this paper, we focused on the learning adaptation method as an enhancement to Online Sequential Extreme Learning Machine (OS-ELM) \cite{4012031} and Constructive Enhancement OS-ELM (CEOS-ELM) \cite{5178608}. We named it as Adaptive OS-ELM (AOS-ELM). The AOS-ELM has capability to handle multiple concept drift problems either changes in the number of attributes (virtual drift/VD) or the number of target classes (real drift/RD) or both at the same time (hybrid drift/HD), also for recurrent context (all concepts occur alternately) or sudden drift (new concept substitutes  previous concepts)  \cite{KunchevaEnsembleOverview08}.  Our scope of attribute changes discussed in this paper is on the feature space concatenation that widely used in data fusion, kernel fusion, and ensemble learning \cite{Dietterich:2000:EMM:648054.743935} and not on the feature selection (irrelevant features removal) methods \cite{Chandrashekar201416}. We compared the performance with nonadaptive sequential ELM: OS-ELM and  CEOS-ELM. We also compared the performance with ELM classifier ensembles as the common adaptive approach for concept drift solution. In the present study, although we focus on the adaptation aspect, we address some possible change detection mechanisms that are suitable for our method. 

A preliminary version of RD and its early results appeared in conference proceedings \cite{6922113}. In this paper, we introduced the new scenarios in VD, HD, and consecutive drifts either recurrent or sudden drift scenarios as well as theoretical background explanation. Our main contributions in this research area can be summarized as follows:

\begin{enumerate}

\item We proposed simple adaptive method as enhancement to OS-ELM and CEOS-ELM for addressing concept drifts issue. Unlike ensemble systems \cite{39313598,Huang201532} that need to manage the complex combination of a vast number of classifiers, we pursue a single classifier for simple implementation while retaining comparable performance for handling multiple (consecutive)  drifts. \item We denote the training data from different $\textstyle{S}$ concepts (sources or contexts), using the symbol $\mathbf{X_{s}}$ for training data and $\mathbf{T_{s}}$ for target data. We used the subscript font without parenthesis to show the source number. 
\item We denote the drift event  using the symbol $\genfrac{}{}{0pt}{}{\ggg}{VD}$, where the subscript font shows the drift type. E.g., the Concept 1 has virtual drift event to be replaced by Concept 2 (Sudden drift) : $\mathbf{C}_1 \genfrac{}{}{0pt}{}{\ggg}{VD} \mathbf{C}_2$. The Concept 1 has real drift event to be replaced by Concept 1 and Concept 2 recurrently  (Recurrent context) in the shuffled composition : $\mathbf{C}_1 \genfrac{}{}{0pt}{}{\ggg}{RD} shuffled(\mathbf{C}_1,\mathbf{C}_2)$. 

\item We introduced a simple unified platform to handle a hybrid drift (HD) when changes in the number of attributes and the number of target classes occurred at the same time. 

\item We elaborated how the AOS-ELM for transfer learning using hybrid drift strategy. Transfer learning focuses on extracting the knowledge from one or more source task domains and applies the knowledge to a different target task domain  \cite{10.1109/TKDE.2009.191}. Concept drift focuses on the  time-varying domain with a small number of current data available. In contrast, transfer learn\item We denote the training data from different $\textstyle{S}$ concepts (sources or contexts), using the symbol $\mathbf{X_{s}}$ for training data and $\mathbf{T_{s}}$ for target data. We used the subscript font without parenthesis to show the source number. 
\item We denote the drift event  using the symbol $\genfrac{}{}{0pt}{}{\ggg}{VD}$, where the subscript font shows the drift type. E.g., the Concept 1 has virtual drift event to be replaced by Concept 2 (Sudden drift) : $\mathbf{C}_1 \genfrac{}{}{0pt}{}{\ggg}{VD} \mathbf{C}_2$. The Concept 1 has real drift event to be replaced by Concept 1 and Concept 2 recurrently  (Recurrent context) in the shuffled composition : $\mathbf{C}_1 \genfrac{}{}{0pt}{}{\ggg}{RD} shuffled(\mathbf{C}_1,\mathbf{C}_2)$. ing is not associated with time and requires the entire training and testing data set \cite{Yang:2009:TLB:1693243.1693247}. The  example of transfer learning by using HD strategy is the transition from different data set sources but still related and has the same purpose. In this paper, we discussed the transfer learning on numeric handwritten MNIST  \cite{lecun-mnisthandwrittendigit-2010} to alpha-numeric handwritten USPS \cite{roweis-usps}  recognition.

\item Naturally, the AOS-ELM handling strategy was based on recurrent context. We devised an AOS-ELM strategy to handle sudden drift scenario by introducing output marginalization method. This method is also applicable for concept drift in a regression problem.  

\item We studied the effect of increasing the number of hidden nodes, which is treated as one of learning parameters, to improve the accuracy (other learning parameters are input weight, bias, activation function, and regularization factor). We proposed the evaluation parameter to predict the accuracy before the training completed. We applied this assessment parameter actually to prevent 'under-fitting' or nonconvergence condition (the model does not fit the data well enough that makes accuracy performance dropped) when any learning parameter changes such as hidden nodes increased. 

\end{enumerate}

This paper is organized as follows. Section \ref{sec1a} explains some issues and challenges in concept drift, the background of ELM, and ELM in sequential learning. Section \ref{sec2a:the_proposed} presents the background theory and  algorithm derivation of the proposed method. 
In section \ref{sec4}, we focus on the empirical experiments to prove the methods and research questions in regression and classification problem. We use artificial and real data set. The artificial data sets are streaming ensemble algorithm (SEA) \cite{conf/kdd/StreetK01} and STAGGER \cite{Kolter:2007:DWM:1314498.1390333}, which are commonly used as benchmark in sequential learning. The real data sets are handwritten recognition data: MNIST for numeric \cite{lecun-mnisthandwrittendigit-2010} and USPS for alpha-numeric classes \cite{roweis-usps}. 
We studied the effect of hidden nodes increase as one of important learning parameter in section \ref{sec44}. Section \ref{sec5a} discusses research challenges and future directions. The conclusion presents some highlights in Section \ref{sec5}.

\section{Related Works}
\label{sec1a}
\subsection{Notations}
\label{sec1a01}

We specify the notations used throughout this article for easier understanding:

\begin{itemize}
\item Matrix is written in uppercase bold (e.g., $\mathbf{X}$). 

\item Vector is written in lowercase bold (e.g., $\mathbf{x}$).

\item The transpose of a matrix $\mathbf{X}$ is written as $\mathbf{X}^{^\mathrm{T}}$. The pseudo-inverse of  a matrix $\mathbf{H}$ is written as $\mathbf{H}^{\dagger}$. 

\item $\textstyle{f}$, $\textstyle{g}$ will be used as non linear differentiable function (activation function), e.g., sigmoid or tanh function.

\item The amount of training data is $\textstyle{N}$. Each input data $\mathbf{x}$ contains some $\textstyle{d}$ attributes. The target has $\textstyle{m}$ number of classes.An input matrix $\mathbf{X}$ can be denoted as $\mathbf{X}_{\textstyle{d}\times\textstyle{N}}$ and the target matrix $\mathbf{T}$ as $\mathbf{T}_{\textstyle{N}\times\textstyle{m}}$.

\item The hidden layer matrix is $\mathbf{H}$. The input weight matrix is  $\mathbf{A}$. The output weight matrix is    $\mathbf{\beta}$. The matrix $\Delta \mathbf{H}$ is the additional block portion of the matrix $\mathbf{H}$. The matrix $\mathbf{K}$ is the auto correlation matrix of $\mathbf{H}^\mathrm{T}\mathbf{H}$. The inverse of matrix $\mathbf{K}$ is  $\mathbf{P}$.

\item $\mathbf{H}$ can be denoted  as $\mathbf{H}_{\textstyle{N}\times\textstyle{L}}$.  $\mathbf{A}$ can be denoted  as $\mathbf{A}_{\textstyle{d}\times\textstyle{L}}$ and $\mathbf{\beta}$ can be denoted as $\mathbf{\beta}_{\textstyle{L}\times\textstyle{m}}$. $\delta \textstyle{L}$ denotes the additional nodes number of $\textstyle{L}$.

\item When the number of training data $\textstyle{N}\rightarrow\infty$, we employed the online sequential learning method by updating model every time each new training pairs $(\mathbf{x}, \mathbf{t})$ are seen. 
$\mathbf{X_{(0)}}$ is the subset of input data at time $k=0$ as the initialization stage. $\mathbf{X_{(1)}}$,$\mathbf{X_{(2)}}$,...,$\mathbf{X_{(k)}}$ are the subset of input data at the next sequential time. Each subset may have different number of quantity. The corresponding label data is presented as  $\mathbf{T} = \left[\mathbf{T_{(0)}},\mathbf{T_{(1)}},\mathbf{T_{(2)}},...,\mathbf{T_{(k)}} \right]$. We used the subscript font with parenthesis to show the sequence number.  

\item We denote the training data from different $\textstyle{S}$ concepts (sources or contexts), using the symbol $\mathbf{X_{s}}$ for training data and $\mathbf{T_{s}}$ for target data. We used the subscript font without parenthesis to show the source number. 

\item We denote the drift event  using the symbol $\genfrac{}{}{0pt}{}{\ggg}{VD}$, where the subscript font shows the drift type. E.g., the Concept 1 has virtual drift event to be replaced by Concept 2 (Sudden drift) : $\mathbf{C}_1 \genfrac{}{}{0pt}{}{\ggg}{VD} \mathbf{C}_2$. The Concept 1 has real drift event to be replaced by Concept 1 and Concept 2 recurrently  (Recurrent context) in the shuffled composition : $\mathbf{C}_1 \genfrac{}{}{0pt}{}{\ggg}{RD} shuffled(\mathbf{C}_1,\mathbf{C}_2)$. 

\end{itemize}

\subsection{Concept Drift Strategies}
\label{sec1a02}
In this section, we briefly explained the various concept drift solution strategies. Gama, \textit{et. al.} \cite{Gama:2014:SCD:2597757.2523813} explained many concept drift methods have been developed, but the terminologies are not well established. 
According to Gama, \textit{et. al.}, the basic concept drift based on Bayesian decision theory in the classification problem for class output $c$ and incoming data $X$ as:

\begin{equation}
\label{eq:bayes}
P(c|\textbf{X}) = P(c) \frac{P(\textbf{X} |c)}{P(\textbf{X})}
\end{equation}

Concept drift  occurred when  $P(c|\textbf{X})$ has changed; e.g., $\exists\textbf{X}:P_{(0)}(\textbf{X},c) \neq P_{(1)}(\textbf{X},c) $, where $P_{(0)}$ and  $P_{(1)}$ are respectively the joint distribution at time $t_{(0)}$ and $t_{(1)}$. Gama, \textit{et. al.} categorized the concept drift types as following:

\begin{enumerate}
\item Real Drift (RD) refers to changes in  $P(c|\textbf{X})$. The change in  $P(c|\textbf{X})$ may be caused by a change in the class boundary (the number of classes) or the class conditional probabilities (likelihood) $P(\textbf{X}|c)$. The number of classes expanded and different class of data may come alternately, known as recurrent context. A drift, where a new conditional probabilities replaces the previous conditional probabilities while the number of class remained the same, is known as sudden drift. Other terms are concept shift or conditional change \cite{Gao07ageneral}.

\item Virtual Drift (VD) refers to the changes in the distribution of the incoming data (e.g. $P(\textbf{X})$ changes). These changes may be due to incomplete or partial feature representation of the current data distribution. The trained model is built with additional data from the same environment without overlapping the true class boundaries. Other terms are feature change \cite{Gao07ageneral}, temporary drift, or sampling shift. 

\end{enumerate} 

Kuncheva \cite{KunchevaEnsembleOverview08,978-3-540-22144-9} explained the various configuration patterns of data sources over time as random noise, random trends (gradual changes), random substitutions (abrupt or sudden changes), and systematic trends (recurring context).  The random noise will simply be filtered out. A gradual drift occurs when many concepts may re-occur alternately in the gradual stage for a certain period. A consecutive drift takes place when many previously active concepts might keep on changing alternately (recurring context) after some time. The sudden drift (abrupt changes or concept substitutions) is the type that at one time, one concept is suddenly replaced by another concept.

\begin{figure}[!t]
%\centering
\includegraphics[width=3.25in]{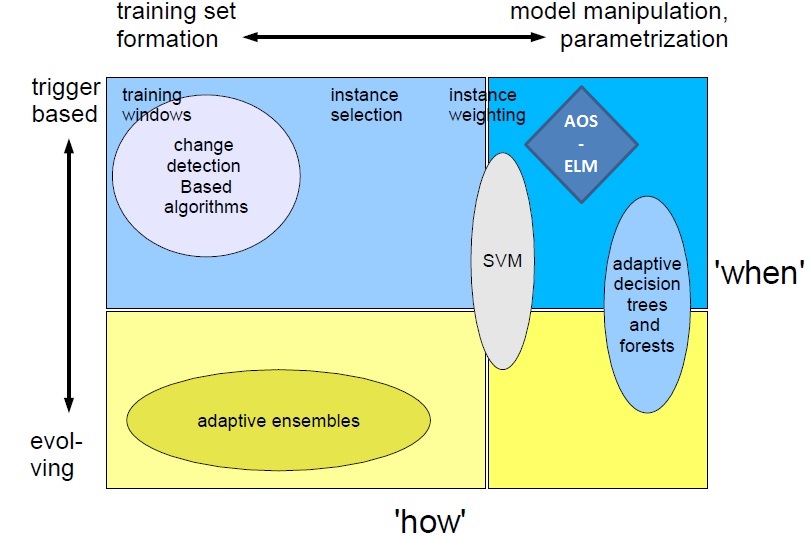}
\caption{The taxonomy quadrant of adaptive supervised learning techniques. Popular concept drift handling methods are indicated by ellipses \cite{39313598}. Our proposed method AOS-ELM is indicated by a dark blue diamond.}
\label{fig:adaptive}
\end{figure}

\v{Z}liobait\.{e} \cite{39313598} proposed a taxonomy of concept drift tackling methods as shown in Fig. \ref{fig:adaptive}. It describes the methods based on when the model is switched on (the 'when' axis) and how the learners adapt to training set formation or design and parametrization of the base learner (The 'how' axis).  The 'when' axis spans drift handling from trigger based to evolving based methods. The 'how' axis spans drift handling from training set formation to model manipulation (or parametrization) methods. 

\v{Z}liobait\.{e} \cite{39313598} explained that most attention on the concept drift tackling methods are drawn to multi-classifier model selection and fusion rules, but little attention on the model construction of base classifier. 

\begin{figure}[!t]
%\centering
\includegraphics[width=3.5in]{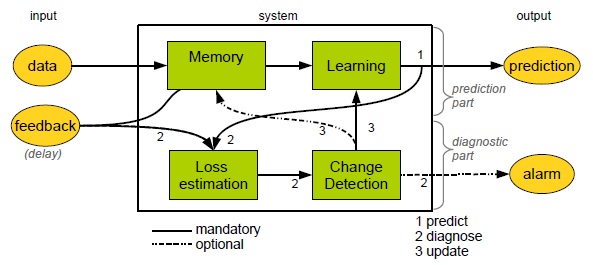}
\caption{A generic scheme for an online adaptive learning algorithm from Gamma, et al. \cite{Gama:2014:SCD:2597757.2523813}. }
\label{fig:adaptive-schema}
\end{figure}

Gama, \textit{et. al.} \cite{Gama:2014:SCD:2597757.2523813} proposed a  complete online adaptive learning scheme that organized four modules: memory, change detection, learning, and loss estimation (See Fig. \ref{fig:adaptive-schema}). These modular components can be integrated, permuted and combined with each other. The key modules are the learning and the change detection modules. Most methods focused on some subset or often mixtures of many types within certain concept drifts.

The learning module refers to the methods for the adaptation strategies of the predictive model. The learning module is categorized based on i) How the model is updated when new data points are available (learning mode): retraining or incremental (online) modes. ii) The behavior of predictive models on time-evolving data (model adaptation): a blind (evolving or implicit) based module or an informed (trigger or explicit) based module. iii) The techniques for maintaining active predictive models (model management): a single model or ensemble model. The change detection module refers to drift detection. The change detection identifies change points or small time intervals when changes occur. 

Each drift employed different solution strategies. The solution for RD is entirely different from VD. If the systematic changes are likely to reappear, we may want to keep past successful classifiers and simply reuse them. If the changes are gradual, we may use a moving window strategy on the training data.  If the changes are abrupt, we can pause the existing static classifiers then retrain the classifier using the new training data. Thus, it is hard to combine simultaneously many strategies at one time to solve many types of concept drift in just a simple platform.

\subsection{ELM in Sequential Learning}
\label{sec20}
In this section, we briefly explained the previous related works of ELM in sequential learning and adaptive environments. 

ELM is getting popularity thanks to its learning speed, generalization capability, and simplicity. Huang \cite{s12559-014-9255-2} explained the term 'Extreme' meant to move beyond conventional artificial neural network learning that required iterative tuning. The ELM moves toward brain like learning in which hidden neurons need not be tuned.

The output function of an SLFN with single hidden layer matrix $\mathbf{H}$ can be presented as the function of: 
\begin{equation}
\label{eq:SLFN}
f_{L}(\mathbf{x})  =  \sum_{i=1}^{L}\mathbf{\beta}_{i}\mathbf{H}(\mathbf{a}_{i},\mathbf{b}_{i},\mathbf{x})
\end{equation}
where $\mathbf{H}^{\dagger}$ is the pseudo inverse of $\mathbf{H}$.

$\mathbf{H}^{\dagger}$ can be  approximated by left pseudoinverse of $\textbf{H}$ as: 

\begin{equation}
\label{eq:Hinvers2}
\hat{\mathbf{\beta}} = (\mathbf{H}^{^\mathrm{T}}\mathbf{H})^{-1}\mathbf{H}^{^\mathrm{T}}\mathbf{T}
\end{equation}
We can use ridge regression or regularized least squares to be: $\hat{\mathbf{\beta}} = \big( \mathbf{H}^{^\mathrm{T}}\mathbf{H}+ \frac{\mathbf{I}}{\textstyle{c}}\big)^{-1}\mathbf{H}^{^\mathrm{T}}\mathbf{T}$. 

Based on [\ref{eq:Hinvers2}], Liang \textit{et.al.} \cite{4012031} proposed online learning for ELM  named OS-ELM.

If we have $\hat{\mathbf{\beta}}_{(0)}$ from  $\mathbf{H}_{(0)}$ filled by the $N_0$ number of training data and $N_1$ incremental batch of data filled  $\mathbf{H}_{(1)}$ , the output weights $\hat{\mathbf{\beta}}_{(1)}$ are approximated by:
\begin{equation}
	\label{eq:oselm}
\hat{\mathbf{\beta}}_{(1)} = \Big ( \left[ \begin{array}{c} \mathbf{H}_{(0)} \\ \mathbf{H}_{(1)} \end{array} \right]^{^\mathrm{T}} \left[ \begin{array}{c} \mathbf{H}_{(0)} \\ \mathbf{H}_{(1)} \end{array} \right] \Big )^{-1} \left[ \begin{array}{c} \mathbf{H}_{(0)} \\ \mathbf{H}_{(1)} \end{array} \right]^{^\mathrm{T}} \left[ \begin{array}{c} \mathbf{T}_{(0)} \\ \mathbf{T}_{(1)} \end{array} \right]
\end{equation}

Both $\textbf{H}_0$ and $\textbf{H}_1$ have a different number of training data but have the same $L$ number of hidden nodes.

If $\mathbf{K} = \mathbf{H}^{^\mathrm{T}}\mathbf{H}$, then we can rewrite: 

\begin{equation}
	\label{eq:oselm1}
	\begin{split}
\hat{\mathbf{\beta}}_{(1)} &=  \mathbf{K}_{(1)}^{-1} \left[ \begin{array}{c} \mathbf{H}_{(0)} \\ \mathbf{H}_{(1)} \end{array} \right]^{^\mathrm{T}} \left[ \begin{array}{c} \mathbf{T}_{(0)} \\ \mathbf{T}_{(1)} \end{array} \right]\\
\end{split}
\end{equation}

The OS-ELM assumes no changes in the number of hidden nodes. However, increasing the number of hidden nodes is required to improve the performance. A CEOS-ELM \cite{5178608}  has addressed this problem by adding hidden nodes in the sequential learning stage. So $ \mathbf{H} =
 \left[ \begin{array}{cc} \mathbf{H}_{(0)} & \Delta\mathbf{H}_{(0)}  \\ \mathbf{H}_{(1)} & \Delta\mathbf{H}_{(1)} \end{array} \right]$. 
 
The sub-matrix $\Delta\mathbf{H}_{(0)}$ is set to a zero block matrix to simplify the computation in accordance with the fact that the previous data is not related to the new hidden nodes. The additional hidden nodes block matrix $\Delta\mathbf{H}_{(1)}$ for $N_{1}$ data, has relation to the additional hidden nodes $\delta L_{(1)}$. 

\noindent Then, we can rewrite $\mathbf{K}_{(1)}$ with $\Delta\mathbf{H}_{(1)}$ as:
\begin{equation}
\label{eq:CEOSELM0}
\begin{split}
\hat{\mathbf{K}}_{(1)} &= \left[ \begin{array}{cc} \mathbf{H}_{(0)} & \mathbf{0}  \\ \mathbf{H}_{(1)} & \Delta\mathbf{H}_{(1)} \end{array} \right]^{^\mathrm{T}} \left[ \begin{array}{cc} \mathbf{H}_{(0)} & \mathbf{0}  \\ \mathbf{H}_{(1)} & \Delta\mathbf{H}_{(1)} \end{array} \right] \\
\end{split}
\end{equation}

If $\hat{\mathbf{P}} =  \hat{\mathbf{K}}^{-1}$ can be solved  using block matrix inversion and Schur complement, then:
\begin{equation}
\label{eq:CEOSELM2}
\begin{split}
\hat{\mathbf{\beta}}_{(1)} &= \hat{\mathbf{P}}_{(1)} \left[ \begin{array}{cc} \mathbf{H}_{(0)} & \mathbf{0}  \\ \mathbf{H}_{(1)} & \Delta\mathbf{H}_{(1)} \end{array} \right]^{^\mathrm{T}}\left[ \begin{array}{c} \mathbf{T}_{(0)} \\ \mathbf{T}_{(1)} \end{array} \right]\\
\end{split}
\end{equation}

It is important to note that both OS-ELM and CEOS-ELM did not address the concept drift issue; e.g., when the number of  attributes $d$  in $\mathbf{X}_{\textstyle{d}\times\textstyle{N}}$ or the number of  classes  $m$ in $\mathbf{T}_{\textstyle{N}\times\textstyle{m}}$  in data set has added. In this paper, we categorized OS-ELM and CEOS-ELM as non-adaptive sequential ELM. 

To the best of our knowledge, no previous single base ELM approach specifically addresses many concept drifts learning \cite{Huang201532}. However, some papers \cite{vanSchaik2015233,1866-9956} already discussed how the ELM implementation in adaptive environment. 

Schaik, \textit{et.al.} \cite{vanSchaik2015233} proposed Online Pseudo Inverse Update Method (OPIUM). OPIUM is based on Greville's method as the incremental solutions to compute the pseudo inverse of matrix.  The pseudo inverse computation can be solved incrementally as linear regression problems and can be adaptive which allows for non stationary data. The derivation of OPIUM is equivalent to the OS-ELM  if the condition  $\textstyle{c}_{(k)} \buildrel \text{d{}ef}\over = (\mathbf{I} - \mathbf{H}_{(k-1)}\mathbf{H}_{(k-1)}^{-1})\mathbf{H}_{(k)} = 0$ met at each iteration.  This condition implies $\mathbf{H}_{(k)}$ is a linear combination of the previous hidden layer $\mathbf{H}_{(k-1)}$ and  the simpler derivation of (\ref{eq:SLFN}) with right pseudo inverse become:
\begin{equation}
\label{eq:Hinvers3}
\hat{\mathbf{\beta}} = \mathbf{T}\mathbf{H}^{\dagger} = \mathbf{T}\mathbf{H}^{^\mathrm{T}}(\mathbf{H}^{^\mathrm{T}}\mathbf{H})^{-1}
\end{equation}

Schaik, \textit{et.al.} defined $\mathbf{\psi}$ as the cross correlation matrix between $\mathbf{T}$ and $\mathbf{H}$ ($\mathbf{T}\mathbf{H}^{^\mathrm{T}}$) and $\mathbf{\theta}$ as the inverse of the auto correlation $\mathbf{H}$ ($\mathbf{H}^{^\mathrm{T}}\mathbf{H})^{-1}$), so $\mathbf{\beta} =  \mathbf{\psi}\mathbf{\theta}$.
According to Greville's method, the solution for $\mathbf{\psi}_{(k)} = \mathbf{\psi}_{(k-1)} + \mathbf{T}_{(k)} \mathbf{H}_{(k)}^{^\mathrm{T}} $. And the solution for  $\mathbf{\theta}_{(k)} = \big(\mathbf{H}_{(k-1)}\mathbf{H}_{(k-1)}^{^\mathrm{T}} +  \mathbf{H}_{(k)} \mathbf{H}_{(k)}^{^\mathrm{T}}\big)^{-1}$, or in short writing as $\mathbf{\theta}_{(k)} = \textstyle{f}(\mathbf{\theta}_{(k-1)},\mathbf{H}_{(k)})$.

Schaik, \textit{et.al.} proposed a simplified version named OPIUM light by computing only the on-diagonal element of $\mathbf{\theta}_{(k)}$. Schaik, \textit{et.al.} applied the OPIUM light for non-stationary data by using different weight $\alpha$ in determining $\mathbf{\beta}_{(k)}$ for the most recent pair $(\mathbf{T}_{(k)},\mathbf{H}_{(k)})$ that appropriate for non-stationary mapping, which are: $\mathbf{\psi}_{(k)} = (2-\alpha) \mathbf{\psi}_{(k-1)} + \alpha\mathbf{T}_{(k)} \mathbf{H}_{(k)}^{^\mathrm{T}} $, and for $\mathbf{\theta}_{(k)} = \big((2-\alpha)\mathbf{H}_{(k-1)}\mathbf{H}_{(k-1)}^{^\mathrm{T}} +  \alpha\mathbf{H}_{(k)} \mathbf{H}_{(k)}^{^\mathrm{T}}\big)^{-1}$.

In our opinion, OPIUM only tackled the real drift case with discriminant function boundary shift in the streaming data (e.g. the frequency shift of sine wave). They implemented the weighting $\alpha$ as a non-stationary mapping parameter between input and output vectors. 

Cao, \textit{et. al.} \cite{1866-9956} proposed two-phase classification algorithm:  First, weighted ensemble classifier based on ELM (WEC-ELM) algorithm, which can dynamically adjust classifier and the weight of training uncertain data to solve the problem of concept drift. Second, an uncertainty classifier based on ELM (UC-ELM) algorithm is designed for the classification of unknown data streams, which considers attribute (tuple) value and its uncertainty, thus improving the efficiency and accuracy. When concept drift occurs, WEC-ELM will dynamically adjust the classifiers and the weight of training data, thus a new classifier will be added to the ensemble until it reached a preset maximum then removed the worst-performing classifier. UC-ELM is designed for the classification of uncertain data streams, which has attributes (tuples) and its uncertainty values. The UC-ELM evaluated uncertainty value for every newly arrived attribute and decided based on the probability of the new attributes belonging to each class, thus improving the efficiency and accuracy. 
In our opinion,  WEC-ELM is categorized as evolving based method by selecting the best-performing classifier, and UC-ELM addressed virtual drift problem by using uncertainty attributes selection. 

\section{Proposed Method}
\label{sec2a:the_proposed}

\subsection{Theoritical Background of AOS-ELM}
\label{sec3}

In sequential learning,  some partial training data arrives in time sequential fashion: $\left \{(\mathbf{x_{(0)}},\mathbf{t_{(0)}}),(\mathbf{x_{(1)}},\mathbf{t_{(1)}}),\cdots,(\mathbf{x_{(k)}},\mathbf{t_{(k)}})\right\}$. 
Learning is the process of constructing function $\hat{\mathbf{\beta}}$ to map between observation and its nature called (class) \cite{ptpr}. When the number of training data $\textstyle{N}\rightarrow\infty$, we need to address the expected value of $\mathbf{\beta_{(\infty)}} = \hat{\mathbf{\beta}}$.

Learning from the data $\mathbf{D_n}$ is the process to select a function $\mathbf{\beta_n}$ from a class of $\mathfrak{B}$ by minimizing of the empirical squared error $e_n(\mathbf{\beta})=\frac{1}{n}\sum_{i=1}^{n}(\mathbf{H_i}\mathbf{\beta}-\mathbf{T_i})^2 $ with the error probability $L(\mathbf{\beta_n})= P\big\{\mathbf{I_{\{\mathbf{H}\mathbf{\beta_n}\}}}\neq \mathbf{T}|\mathbf{D_n} \big\}$ of the resulting classifier. According to \cite{ptpr}, the empirical squared error minimization is consistent under general conditions.

\begin{theorem}
\label{theorem301} 
Assume that $\mathfrak{B}$ is a totally bounded class of functions. If $\mathbf{\beta_n} \in \mathfrak{B} $, then the classification rule obtained by minimizing the empirical squared error over $\mathfrak{B}$ is strongly consistent, that is,
\begin{equation}
\label{eq:consistent0}
P\big\{\lim_{n\to\infty} L(\mathbf{\beta_n}) = L^{*} \big\} \rightarrow 1
\end{equation}
\end{theorem}

Based on Law of Large Numbers (LLN) theorem \cite{prob:2003}  and Theorem \ref{theorem301}, in sequential learning with the number of training data $\textstyle{N}\rightarrow\infty$, we can make sure the consistency of expected value of learning model is $\hat{\mathbf{\beta}} = \mathbf{H}^{\dagger}\mathbf{T}$.

The Concept drift refers to an online supervised learning model when the relation between the input data and the target variable changes over time \cite{Gama:2014:SCD:2597757.2523813}. If the learning model from Concept 1 $\hat{\mathbf{\beta_1}} \in \mathfrak{B_1}$ is bounded by hypothesis space $\mathbb{R}^{\textstyle{m_1}}$ and feature space $\mathbb{R}^{\textstyle{d_1}}$. And the learning model from Concept 2 $\hat{\mathbf{\beta_2}} \in \mathfrak{B_2}$ is bounded by hypothesis space $\mathbb{R}^{\textstyle{m_2}}$ and feature space $\mathbb{R}^{\textstyle{d_2}}$. We defined the real drift as when the  hypothesis space $\mathbb{R}^{\textstyle{m_1}}$ has changed to $\mathbb{R}^{\textstyle{m_2}}$. We scoped the definition for  $\textstyle{m_2} > \textstyle{m_1}$ dimension changes. The virtual drift is when the feature space $\mathbb{R}^{\textstyle{d_1}}$ has changed to $\mathbb{R}^{\textstyle{d_2}}$. We scoped the definition for  $\textstyle{d_2} > \textstyle{d_1}$ dimension changes.

To achieve the consistency of minimized square error in the new hypothesis space or new feature space, the learning model needs a transition map from the former space to the new space. The learning model $\hat{\mathbf{\beta_1}}$ needs a transition space  before it converges to the new learning model $\hat{\mathbf{\beta_2}} \in \mathfrak{B_2} \subset \mathbb{R}^{\textstyle{m_2}}$ . Our transition space idea was inspired by geometric approach for solving many problems in the fields of pattern recognition and machine learning \cite{doi:10.1137/1034101, Strack:2013:GAS:2519716}. 

For transition space, we propose two approaches: i) Assign the random  coordinates in the new concept space. ii) Assign the equivalent projection coordinates in the new design space. The first approach is suitable for VD scenario, in which we assigned the new random coordinates as the new input weight parameters. The second approach is suitable for RD situation, by setting the equivalent projection coordinates in the new space (e.g. The $(X_1)$ in 1-D coordinate has corresponding 2-D projection coordinates as $(X_1,0)$). 

Here, we relate the ELM theory to the context of AOS-ELM concept drift scenarios as follows: 

\begin{enumerate}
\item Scenario 1: virtual drift (VD).

Huang, \textit{et. al.} \cite{Huang201532} explained interpolation theory from ELM point of view  as stated by the following description:
\begin{theorem}
\label{theorem31}
Given any small positive value $\epsilon > 0$, any activation function which is infinitely differentiable in any interval, and $N$ arbitrary distinct samples $(\mathbf{x}_i, \mathbf{t}_i) \in \mathbf{R}^d \times \mathbb{R}^m$, there exists $L < N$ such that for any input weight and bias pair $\{\mathbf{a}_i, \mathbf{b}_i\}^{L}_{i=1} $ randomly generated from any interval of $\mathbf{R}^d \times \mathbb{R}$, according to any continuous probability distribution, then with probability one, $\|\mathbf{H}\mathbf{\beta} - \mathbf{T} \|< \epsilon$. Furthermore, if $L = N$, then with probability one, $\|\mathbf{H}\mathbf{\beta} - \mathbf{T} \| = 0$. 
\end{theorem}

According to  Theorem \ref{theorem31} and Learning Principle I of ELM Theory \cite{s12559-014-9255-2}, the input weight  and bias as hidden nodes $\mathbf{H}$ parameters are independent of training samples and their learning environment through randomization. Their independence is not only in initial training but also in any sequential training stages.  Thus, we can adjust the input weight and bias pair $\{\mathbf{a}_i, \mathbf{b}_i\}^{L}_{i=1} $ on any sequential  stages and still make sure with probability one that $\|\mathbf{H}\mathbf{\beta} - \mathbf{T} \|< \epsilon$.

\item Scenario 2: real drift (RD).

Huang, \textit{et. al.} \cite{Huang201532} explained universal approximation capability of ELM as described by the following theorem:

\begin{theorem}
\label{theorem32}
Given any nonconstant piecewise continuous function $\textstyle{g} :
\mathbb{R}_d \rightarrow \mathbb{R}$, if span $\{\textstyle{g}(\mathbf{a}, \mathbf{b}, \mathbf{x}) : (\mathbf{a}, \mathbf{b})  \mathbb{R}_d \times \mathbb{R}\}$ is dense in L2, for any continuous target function $\textstyle{f}$ and any function sequence $\{\textstyle{g}(\mathbf{a}_i, \mathbf{b}_i, \mathbf{x})\}^{L}_{i=1}$ randomly generated according to any continuous sampling distribution, $\lim_{L\rightarrow\infty}\|\textstyle{f} {-} \textstyle{f}_L\|= 0$ holds with probability one if the output weights $\mathbf{\beta}_i$ are determined by ordinary least square to minimize $\|\textstyle{f}\mathbf{x}) {-} \Sigma^{L}_{i=1}\mathbf{\beta}_i \textstyle{g}(\mathbf{a}_i, \mathbf{b}_i, \mathbf{x})\|$.
\end{theorem}

Based on Theorem \ref{theorem32} and inspired by the related works \cite{5178608,6922113}, we devised the AOS-ELM real drift  capability by modifying the output matrix with zero block matrix concatenation to change the size dimension of the matrix without changing the value. Zero block matrix has meant the previous $\mathbf{\beta}_{(k-1)}$ has no knowledge about the new concept. ELM can approximate any complex decision boundary,  as long as the output weights $\mathbf{\beta}_i$ are determined by ordinary least square to keep the minimum. 

\end{enumerate}

\subsection{AOS-ELM Algorithms}
\label{sec30}

In this section, we presented the AOS-ELM pseudo codes  \footnote{The Matlab source code, data set and demo file implementation are available at https://github.com/abudiman250172/adaptive-OS-ELM.}  in the $k^{th}$ sequential with $\mathbf{X}_{(k)}$ training input and $\mathbf{T}_{(k)}$ target to update  $Model_{(k)}$. 

Basically, we have three pseudo codes, namely OSELMSeq (Algorithm \ref{oselmseq}) as OS-ELM and CEOS-ELM pseudo codes; the AOSELMVDSeq (Algorithm 1 as  AOS-ELM pseudo codes tackling virtual drift; and AOSELMRDSeq (Algorithm 2 as AOS-ELM pseudo codes for addressing real drift. We can combine each pseudo code together to form a hybrid drift Algorithm. We can increase  the hidden nodes using CEOS-ELM in Algorithm 3 after AOSELMVDSeq or AOSELMRDSeq. For initialization, basically we can use any ordinary ELM initialization in offline  learning mode. 

\begin{figure}[!t]
\centering
\includegraphics[width=3in]{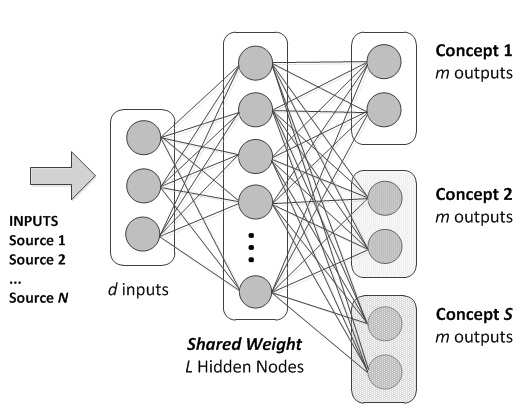}
\caption{Output marginalization in AOS-ELM. The new block of output nodes assembled when the new concept $\textstyle{S}$ presented. Each concept has the same $m$ output nodes quantity. Total output nodes becomes $\textstyle{S} \times m$ output nodes.}
\label{fig:SEA}
\end{figure}

For sudden drift scenario, we proposed output marginalization method by adding the new output nodes when the new concept presented (See Fig. \ref{fig:SEA}) and marginalized the output result by defining the  $\mathbf{Y_s}$ class of concept S is  $= {\underset  {y_s}{\operatorname {arg\,max}}}\,\mathbf{T}(y_s)$. We scoped the new concept has the same output nodes quantity with the previous concept. 
Output marginalization is by shifting the ELM output to the output nodes that belonging to the new concept and ignoring the previous concept output nodes. This strategy is similar with classifier pruning in ELM ensemble. However, in output marginalization, we can reactivate the previous concepts by shifting back to the previous output nodes. If we want to forget the last concept totally, we can quickly delete the previous output nodes without impacting the generalization performance, or we can increase the hidden nodes at the same time with the drift event.

In regression, because we have only one output node, then we can employ sudden drift scenario by  amplifying the related output node of the concept with a constant value that makes the maximum output $\mathbf{Y_s}$ approximated to 1. 

The systematic rules make AOS-ELM more flexibe to handle complex consecutive drifts scenario. The AOS-ELM only stored the previous output weight $\mathbf{\beta}_{L \times m}$ and auto correlation $\mathbf{K}_{L \times L}$. The auto correlation $\mathbf{K}$ did not keep the training data. This makes AOS-ELM scalable for big streaming data without impacting the computation performance. 

To improve the accuracy, we define the target values $\in \{0,1\}$, so that $\mathbf{Y}$ class is  $= {\underset  {y}{\operatorname {arg\,max}}}\,\mathbf{T}(y)$. According to \cite{Iosifidis2015158}, the target values $\in \{0,1\}$ is equivalent with  $\in \{-1,1\}$. 

\begin{algorithm}
\caption{Algorithm OSELMSeq  \{OS-ELM Sequential \}  }

% \caption{OSELMSeq($\mathbf{X}_{(k)},\mathbf{T}_{(k)}),\mathbf{A}_{(k)},\mathbf{b}_{L},\mathbf{K}_{(k-1)})$,\\
% $\mathbf{\beta}_{(k-1)},
% \textstyle{IncreaseHiddenNodes}$) }
\label{oselmseq}
\begin{algorithmic}[1]
\REQUIRE $\mathbf{X}_{(k)}  \in \left[-1,1\right] \mathbb{R}^{d \times N}$,
$\mathbf{T}_{(k)} \in \left[0,1\right] \mathbb{R}^{N \times m}$, \\
$\mathbf{A}_{(k)}$,
$\mathbf{b}_{L}$, 
$\mathbf{K}_{(k-1)}$,
$\mathbf{\beta}_{(k-1)}$
\ENSURE $\mathbf{\beta}_{(k)}$,$\mathbf{K}_{(k)}$
\STATE Compute $\mathbf{H}_{(k)} = \textstyle{g}(\mathbf{A}_{(k)} \cdot \mathbf{X}_{(k)} + \mathbf{b}_{L})$ \\
\IF {$\textstyle{IncreaseHiddenNodes} == \TRUE$}
    \STATE $\Delta \mathbf{A}_{d \times \delta L} =  \textstyle{Random Numbers}(\left[-1,1\right],\mathbb{R}^{d \times \delta L} ) $ 
    \STATE $\Delta \mathbf{b}_{\delta L} =  \textstyle{Random Numbers}(\left[-1,1\right],\mathbb{R}^{\delta L} ) $ 
    \STATE $\mathbf{A}_{(k)} = \begin{bmatrix} \mathbf{A}_{(k)}
        \Delta \mathbf{A}_{d \times \delta L} \end{bmatrix}$ \\
    \STATE $\mathbf{b}_{L} = \begin{bmatrix} \mathbf{b}_{(L)}
        \Delta \mathbf{b}_{\delta L} \end{bmatrix}$ \\
    \STATE Compute $\Delta \mathbf{H}_{(k)} = \textstyle{g}(\Delta \mathbf{A}_{d \times \delta L} \cdot \mathbf{X}_{(k)} + \Delta \mathbf{b}_{\delta L})$ \\
    % \COMMENT {The hidden nodes increase $\Delta \mathbf{H}$ on  CEOS-ELM.}
%   \STATE $\mathbf{H}_{(k)} = \begin{bmatrix} \mathbf{H}_{(k)}
%         \Delta \mathbf{H}_{(k)} \end{bmatrix}$ \\ 
    \STATE Compute $\mathbf{K}_{(k)} = \textstyle{f} \big( \mathbf{K}_{(k-1)}, \mathbf{H}_{(k)}, \Delta\mathbf{H}_{(k)} \big)$\\
    \STATE Compute $\mathbf{\beta}_{(k)} = \textstyle{f} \big(\mathbf{\beta}_{(k-1)}, \mathbf{K}_{(k)}, \mathbf{H}_{(k)}, \Delta\mathbf{H}_{(k)},\mathbf{T}_{(k)} \big)$\\
    \COMMENT{Using CEOS-ELM Method}
   
\ELSE
    \STATE Compute $\mathbf{K}_{(k)} = \mathbf{K}_{(k-1)} +  \mathbf{H}_{(k)}^{^\mathrm{T}} \mathbf{H}_{(k)}$
    \STATE Compute $\mathbf{\beta}_{(k)} = \textstyle{f} \big( \mathbf{\beta}_{(k-1)}, \mathbf{K}_{(k)},\mathbf{H}_{(k)}, \mathbf{T}_{(k)} \big)$
\ENDIF
\RETURN $\mathbf{\beta}_{(k)}$,$\mathbf{K}_{(k)}$\\
\end{algorithmic}
\end{algorithm}

\section{Experiments}
\label{sec4}

\subsection{Experiments Design in Classification}
\label{sec41}

To verify our method, we designed some experiments with the following purposes: 
\begin{itemize}
\item To investigate the effectiveness of AOS-ELM on tackling three concept drift scenarios (VD, RD, HD) in two sequential patterns (sudden changes, recurring context). We used various data set starting with synthetic data set (SEA, STAGGER) then with real data set in handwritten recognition (MNIST, USPS).  Each data set has different drift characteristics. This experiment is presented in Section \ref{sec42} and \ref{sec43}. We also demonstrated the AOS-ELM capability as drift detection role in section \ref{sec42a} using SEA data set.
\item To investigate the effectiveness of AOS-ELM on transfer learning to combine different data set sources.  This experiment is presented in Section  \ref{sec43} using two data set sources (MNIST, and USPS)  in handwritten recognition problem.  
\item To investigate the effect of hidden nodes increase in the drift events and how it impacts performance. This experiment is presented in section \ref{sec44}.
\end{itemize}

We used Matlab \texttrademark{} running on Microsoft Windows \texttrademark{} Computer with 4 cores 2.5 GHz processor and 8 GB memory.

Our experiments are organized as follows:
\begin{enumerate}
\item Simulation benchmark tests on the datasets that commonly used in concept drift handling of stream data, e.g. SEA \cite{conf/kdd/StreetK01} and STAGGER \cite{Kolter:2007:DWM:1314498.1390333} (See Table \ref{table:dataset}). Both datasets are binary classification problem. SEA has 3 inputs with random integer values from 0 to 9.  STAGGER has three inputs with multiple category values from 1 to 3 (Total inputs are 9). SEA and STAGGER are the examples of concept drift that caused by discriminant function changes while the number of attributes and classes from all concepts are still same. The change type is sudden drift. The expected result is the classifier has good performance for the newest concept \cite{978-3-540-22144-9}. 

\item We tested our algorithm with real-world public data sets from MNIST numeric (0 to 9) \cite{lecun-mnisthandwrittendigit-2010} and the USPS alphanumeric (A to Z, 0 to 9) handwritten dataset \cite{roweis-usps}. We used original grey-level image attributes [$X_{grey}$] of MNIST data set and the combination of [$X_{grey}$] with  additional attributes from the 9x9 bins histogram of orientated gradients ($X_{HOG}$) of grey-level image features \cite{5309700}. For USPS, we added more data with Gaussian random and salt-pepper noises. Refer to Table \ref{table:dataset} for detail  data set information. 

\item We designed the initial input weights and bias based on robust OS-ELM with regularization scalar $c$ (ROS) \cite{Hoang:2007:ROS:1418707.1418840} and then based on initial random from the normal distribution (NORM). The activation function is sigmoid. The pseudo inverse function is the orthogonal projection using ridge regularization.

\item Let's define the following concept as :
\begin{itemize}
\item $\mathbf{C_1}$ is  MNIST$[X_{grey}]$ class (1-6);
\item $\mathbf{C_2}$ is  MNIST${[X_{grey}]}$  class (7-10);
\item $\mathbf{C_3}$ is  MNIST${[X_{grey}X_{HOG}]}$  class (1-6);
\item $\mathbf{C_4}$ is  MNIST${[X_{grey}X_{HOG}]}$  class (7-10);
\item $\mathbf{C_5}$ is  MNIST${[X_{grey}X_{HOG}]}$  class (1-10);
\item $\mathbf{C_6}$ is USPS${[X_{grey}X_{HOG}]}$  class (1-10, A-Z);
\end{itemize}

We followed the simulated concept drift methods in Dries, \textit{et.al} \cite{Dries:2009:ACD:1672752.1672754}. 
We simulated sudden drift by splitting the composition into two groups, e.g., $\mathbf{C_1}$ and $\mathbf{C_2}$ and recurring context by shuffled the composition of $\mathbf{C_1}$ and $\mathbf{C_2}$. 
We set the sequential training flow to be the following drift equation: 

We followed the simulated concept drift methods in Dries, \textit{et.al} \cite{Dries:2009:ACD:1672752.1672754}. 
We simulated sudden drift by splitting the composition into two groups, e.g., $\mathbf{C_1}$ and $\mathbf{C_2}$ and recurring context by shuffled the composition of $\mathbf{C_1}$ and $\mathbf{C_2}$. 
We set the sequential training flow to be the following drift equation: 

\begin{enumerate}
\item Experiment 1 - Virtual drift: 

MNIST$[X_{grey}]$ $\genfrac{}{}{0pt}{}{\ggg}{VD}$ MNIST$[X_{grey}X_{HOG}]$
\item Experiment 2 - Real Drift:

For recurring context:
$\mathbf{C_1} \genfrac{}{}{0pt}{}{\ggg}{RD} shuffled (\mathbf{C_1},\mathbf{C_2})$ 

For sudden drift:
$\mathbf{C_1} \genfrac{}{}{0pt}{}{\ggg}{RD} \mathbf{C_2}$ 

\item Experiment 3 - Hybrid Drift:

$\mathbf{C_1} \genfrac{}{}{0pt}{}{\ggg}{HD} shuffled (\mathbf{C_3},\mathbf{C_4})$ 

\item Experiment 4 - MNIST+USPS Transfer Learning: 

$\mathbf{C_5} \genfrac{}{}{0pt}{}{\ggg}{RD} \mathbf{C_6}$
\end{enumerate}

\item We measured the performance based on Table \ref{table:mea}. The testing accuracy and Cohen's Kappa are to show the quantitative measurement. The predictive accuracy is to demonstrate the trend in a line chart. The sudden drift performance is based on the forgetting capability that compared the testing accuracy of the latest concept against all the previous concepts.

\item We compared the AOS-ELM performance with non-adaptive online sequential and offline version of ELM classifier. The performance expectation of sequential version classifier is to approximate the offline version of the classifier (desiderata for online classifiers \cite{978-3-540-22144-9}). We also compared with adaptive ELM ensemble method (See Fig. \ref{fig:mcs-drift}). We designed the hierarchical ensemble using two models of ELM classifier with different roles (See Fig. \ref{fig:mcs-drift}). The first role is a binary classifier that acts as a director based on one against all (OAA) classification. The binary classifier needs all sequential training data to be recalled (full memory). Another role is the data classifier. This ensemble requires total  $2S-1$ classifiers for  $\textstyle{S}$ concepts, thus not effective for consecutive concept  drift case e.g. SEA concepts. The ensemble also applied outdated classifier pruning when the ensemble detects the previous attributes need to be replaced.

\begin{figure}[!t]
\centering
\includegraphics[width=3.25in]{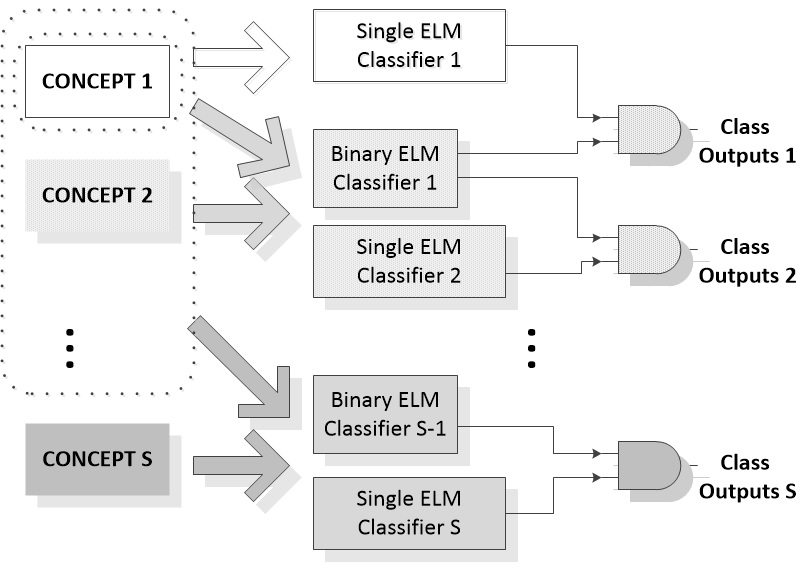}
\caption{Hierarchical ELM ensemble for MNIST+USPS Experiment. The gray shadow  showed the new classifiers assembled when the new concept  presented.}
\label{fig:mcs-drift}
\end{figure}

\end{enumerate}

\begin{table}[!t] 
\small
\label{alltable1}
\renewcommand{\arraystretch}{1.3}
    \caption{Concept Drift Scenarios, Compared Methods and Sequential Patterns.}
    %\footnotesize
    \centering
    \subfloat[The Experiment design scenarios\label{table:design}]{% 
%\begin{tabular}{|p{2.0cm}|p{0.35cm}|p{0.35cm}|p{0.35cm}|p{2.5cm}|}
\begin{tabular}{|l|c|c|c|l|}
\hline
%\noalign{\smallskip}
Data Set & VD & RD & HD & Compared\\
 &  &  &  & Methods\\
%\noalign{\smallskip}
\hline
\hline
%\noalign{\smallskip}
SEA  &  \textbf{-} &  \checkmark &  \textbf{-} & OS-ELM, CEOS-ELM, Kolter \cite{Kolter:2007:DWM:1314498.1390333}\\
\hline
STAGGER  &  \textbf{-} &  \checkmark &  \textbf{-} & OS-ELM, CEOS-ELM, Kolter \cite{Kolter:2007:DWM:1314498.1390333}\\
\hline
MNIST &  \checkmark   & \checkmark   & \checkmark & OS-ELM, Offline ELM, ELM Ensemble\\
\hline
MNIST+USPS & \textbf{-}  &  \checkmark &   \checkmark & OS-ELM, Offline ELM, ELM Ensemble\\
\hline
\end{tabular}
}
	\\
    \subfloat[Concept Drift Sequential Patterns\label{table:flow}]{%
\centering     
%\begin{tabular}{|m{1.75cm}|m{2.25cm}|c|}
\begin{tabular}{|l|l|c|}
\hline
%\noalign{\smallskip}
Data Set & Sequential Patterns Scenarios  & Cause of shift\\
%\noalign{\smallskip}
\hline
\hline
%\noalign{\smallskip}
SEA  &  Sudden change  & Linear discriminant function  \\
\hline
STAGGER  &  Sudden change & Logical discriminant rule \\
\hline
MNIST &  Sudden change , Recurring Context & Additional attributes or classes  \\
\hline
USPS & Recurring Context & Additional attributes or classes\\
\hline
\end{tabular}
}
  \end{table}

\begin{table}[!t]
\label{alltable2}
\renewcommand{\arraystretch}{1.3}
    \caption{Data set Dimension, Quantity, Evaluation method, and Performance Measurement.}
    %\footnotesize
    \centering
    \subfloat[Data Set dimension and  Quantity\label{table:dataset}]{%
\centering     
%\begin{tabular}{|p{1.5cm}|p{1cm}|p{1cm}|p{1cm}|p{1.5cm}|}
\begin{tabular}{|l|l|l|l|l|}
\hline
%\noalign{\smallskip}
Data Set & Concepts & Inputs & Outputs & Quantity \hspace{0.5cm} ($\times$ Concepts)   \\
%\noalign{\smallskip}
\hline
\hline
%\noalign{\smallskip}
SEA  &  4   &  3 & 2   &  20000 ($\times$4)  \\
\hline
STAGGER  &  3 & 9 & 2   &  4400 ($\times$3)  \\
\hline
MNIST &  2   & 784, 865 & 10     & 70000 ($\times$2)   \\
\hline
USPS & 1 & 865 & 36  & 48908 ($\times$1) \\
\hline
\end{tabular}
    }
    \\
    \subfloat[Evaluation  Method\label{table:eval}]{%
\centering     
%\begin{tabular}{|p{1.25cm}|p{2.75cm}|p{1.35cm}|p{1.35cm}|}
\begin{tabular}{|l|l|l|l|}
\hline
%\noalign{\smallskip}
Data Set & Evaluation Method & Training & Testing \\
%\noalign{\smallskip}
\hline
\hline
%\noalign{\smallskip}
SEA  &  5-Fold Cross Validation      &  16000 ($\times$4) & 4000 ($\times$4) \\
\hline
STAGGER  & 5-Fold Cross Validation    &  3520 ($\times$3) & 880 ($\times$3) \\
\hline
MNIST &  Holdout ($10\times$ trials)   & 60000 ($\times$2) & 10000 ($\times$2)  \\
\hline
USPS & Holdout ($10\times$ trials) &  35050 & 13858 \\
\hline
\end{tabular}
}
    \\
    \subfloat[Performance Measurements\label{table:mea}]{%
\centering     
%\begin{tabular}{|p{2.75cm}|p{5cm}|}
\begin{tabular}{|p{2.75cm}|p{5cm}|}
\hline
%\noalign{\smallskip}
Measure & Specification \\
%\noalign{\smallskip}
\hline
\hline
%\noalign{\smallskip}
Accuracy &  The accuracy of classification in \% from $\frac{\#\textit{Correctly Classified}}{\#\textit{Total Instances}} $ \\
\hline
Predictive Accuracy &  The accuracy measurement of the future sequential training data  \cite{Kolter:2007:DWM:1314498.1390333}. 
% in sequential learning (using the next batch of training data for validation)   
% We measured based on the output error (OE) $E(\textbf{H}) =
% \|\textbf{H}\beta - \textbf{T}\|$ for predictive accuracy \cite{5178608}.   
\\
\hline
Testing Accuracy &  The accuracy measurement of the testing data set which excluded from the training. \\
\hline
Forgetting capability &  The testing accuracy differences between the current concept with the previous concepts. \\
\hline
Cohen's Kappa and kappa error  &  The  statistic measurement of inter-rater agreement for categorical items. \\
% We used unweighted Cohen's kappa and kappa error \cite{Cardillo:2007:Online}.   \\
\hline
\end{tabular}
    }
  \end{table}

\subsection{SEA and STAGGER Concepts Result}
\label{sec42}

We addressed the question whether non-adaptive OS-ELM and CEOS-ELM with $\delta L$ increase could handle the concept drift situation. 
We compared between AOS-ELM with no $\delta L$ increase (AOS-ELM1) and with $\delta L$ increase (AOS-ELM2). We used 5-fold cross-validation and compared between NORM and ROS parameter. For SEA, parameters: $L_0= 3000$  and $\delta L=500$  increase per drift. For STAGGER, parameters: $L_0= 9$  and $\delta L=5$ hidden nodes increase per drift.  

\begin{figure}[!t]
\centering
\begin{tabular}{c}
\subfloat[The SEA concept]{\includegraphics[width=3.25in]{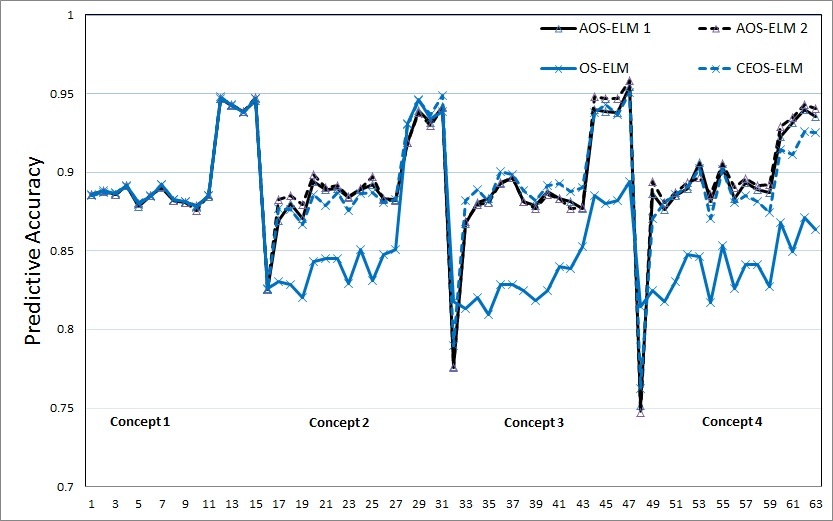}} \\ 
\subfloat[The STAGGER concept]{\includegraphics[width=3.25in]{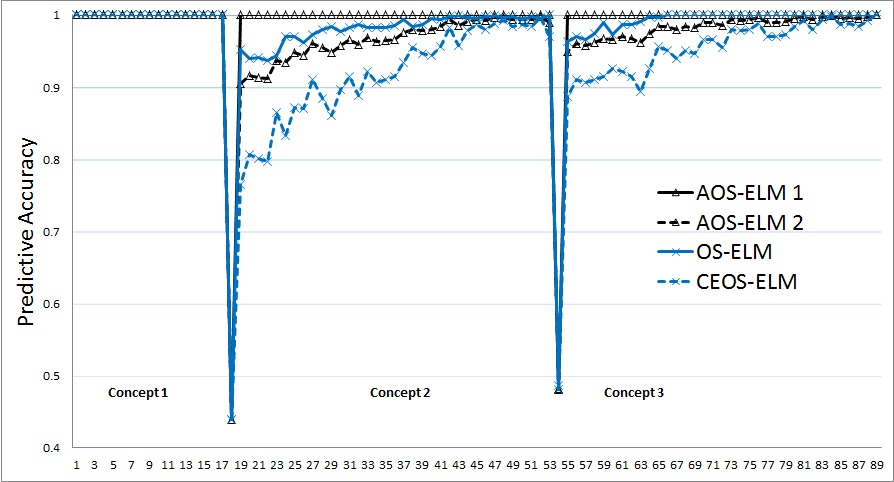}} 
\end{tabular}
\caption{Predictive accuracy of AOS-ELM1 (black line $\triangle$) and AOS-ELM2 (black dots $\triangle$) compared with OS-ELM (blue line $\times$) and CEOS-ELM (blue dots $\times$).}
\label{fig:sea-elm}
\end{figure}

The AOS-ELM has better accuracy with better recovery time (See Table \ref{table:441},\ref{table:442}) than CEOS-ELM, whereas non-adaptive OS-ELM fails (See Fig. \ref{fig:sea-elm}). The AOS-ELM2  improved the forgetting capability better than AOS-ELM1. In comparison with Kolter, \textit{et.al} result using dynamically weighted majority (DWM) of naive Bayes (DWM-NB) for SEA, AOS-ELM result is near to the DWM result. Comparison with inducing decision trees (DWM-ITI) for STAGGER \cite{Kolter:2007:DWM:1314498.1390333}, AOS-ELM outperformed DWM. (See Table \ref{table:441} and \ref{table:442}).  

\begin{table}[!t]
    \caption{Average testing accuracy in \% for each concept between OS-ELM, CEOS-ELM, AOS-ELM1 and AOS-ELM2.}
    % \subref{table:unbalance_mnist} split composition and \subref{table:unbalance_mnist2} split and shuffled composition}
    \footnotesize
    \centering
    \subfloat[Testing accuracy in \% for SEA with $\mathbf{C}_1 \genfrac{}{}{0pt}{}{\ggg}{RD} \mathbf{C}_2 \genfrac{}{}{0pt}{}{\ggg}{RD} \mathbf{C}_3 \genfrac{}{}{0pt}{}{\ggg}{RD} \mathbf{C}_4 $.\label{table:441}]{% 
      
\begin{tabular}{ |p{0.75cm}|p{1cm}|p{1cm}|p{1cm}|p{1cm}|p{1cm}| } 
\hline
Method & Parameter & $\mathbf{C}_1$ & $\mathbf{C}_2$ & $\mathbf{C}_3$ & $\mathbf{C}_4$\\
\hline
\hline
\multirow{2}{4em}{OS-\\ELM} & NORM & $89.48\pm0.33$ & $86.47\pm0.37$ & $83.41\pm0.93$ & $85.32\pm0.45$\\\cline{2-6} 
& ROS & $89.51\pm0.37$ & $86.43\pm0.38$ & $83.49\pm0.95$ & $85.31\pm0.49$  \\\cline{1-6}

\multirow{2}{4em}{CEOS-ELM} & NORM & $82.40\pm0.27$ & $89.33\pm0.43$ & $75.80\pm0.87$ & $90.30\pm0.35$ \\\cline{2-6} 
& ROS & $84.54\pm0.59$ & $89.96\pm0.59$ & $77.97\pm1.05$ & $90.26\pm0.69$  \\\cline{1-6}
\multirow{2}{4em}{\textbf{AOS-ELM1}} & NORM & $89.84\pm0.28$ & $90.03\pm0.25$ & $89.67\pm0.61$ & $\mathbf{90.33}\pm0.39$ \\\cline{2-6} 
& ROS & $89.76\pm0.34$ & $90.02\pm0.25$ & $89.76\pm0.55$ & $\mathbf{90.34}\pm0.38$  \\\cline{1-6}

\multirow{2}{4em}{\textbf{AOS-ELM2}} & NORM & $50.58\pm1.18$ & $50.71\pm1.19$ & $48.50\pm10.10$ & $\mathbf{90.34}\pm0.30$ \\\cline{2-6} 
& ROS & $65.78\pm1.21$ & $65.67\pm1.19$ & $64.09\pm1.89$ & $\mathbf{90.14}\pm1.34 $ \\\cline{1-6}

\end{tabular}
    }
    \\
    \subfloat[Testing accuracy in \% for STAGGER with $\mathbf{C}_1 \genfrac{}{}{0pt}{}{\ggg}{RD} \mathbf{C}_2 \genfrac{}{}{0pt}{}{\ggg}{RD} \mathbf{C}_3$.\label{table:442}]{%
     
\begin{tabular}{ |p{0.8cm}|p{1cm}|c|c|c| } 
\hline
Method & Parameter & $\mathbf{C}_1$ & $\mathbf{C}_2$ & $\mathbf{C}_3$ \\
\hline
\hline
\multirow{2}{4em}{OS-\\ELM} & NORM & $51.89\pm3.48$ & $81.61\pm4.74$ & $67.18\pm5.80$ \\\cline{2-5} 
& ROS & $49.77\pm1.96$ & $84.16\pm1.61$ & $66.93\pm2.00$ \\\cline{1-5} 
\multirow{2}{4em}{CEOS-ELM} & NORM & $21.98\pm1.57$ & $53.66\pm4.34$ & $97.84\pm4.32$ \\\cline{2-5} & ROS & $23.23\pm1.93$ & $52.11\pm2.35$ & $99.27\pm1.45$  \\\cline{1-5}
\multirow{2}{4em}{\textbf{AOS-ELM1}} & NORM & $97.64\pm1.95$ & $\mathbf{100.00}\pm0.00$ & $\mathbf{100.00}\pm0.00$ \\\cline{2-5} 
& ROS & $\mathbf{100.00}\pm0.00$ & $\mathbf{100.00}\pm0.00$ & $\mathbf{100.00}\pm0.00$  \\\cline{1-5}
\multirow{2}{4em}{\textbf{AOS-ELM2}} & NORM & $59.66\pm5.65$ & $70.91\pm10.93$ & $\mathbf{100.00}\pm0.00$ \\\cline{2-5} 
& ROS & $56.20\pm9.56$ & $69.41\pm14.05$ & $\mathbf{100.00}\pm0.00$  \\\cline{1-5}
\end{tabular}
    }
  \end{table}

\subsection{Concept Drift Detection}
\label{sec42a}

\begin{figure}[!t]

	\centering
		\centerline{\includegraphics[width=3.25in]{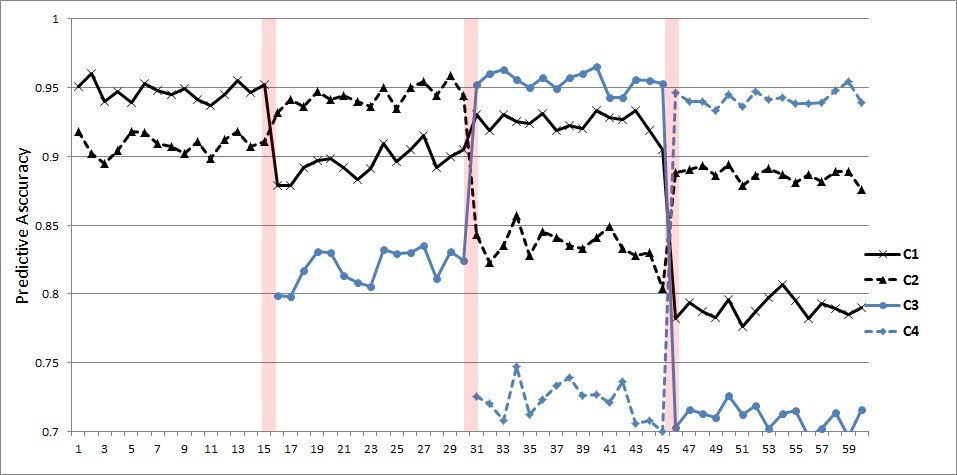}}
		\caption{Predictive accuracy of AOS-ELM in SEA for each concept with \textit{m} output (See Fig. \ref{fig:SEA}). We can consider the intersection point between accuracy decrease of previous concept and accuracy increase of current concept as change point (displayed as thin vertical shadow line).}
	\label{fig:drift}
\end{figure}

The drift detection works based on loss estimation (See Fig. \ref{fig:adaptive-schema}) that compared current prediction accuracy with the previous feedback. 
Using similar method on \cite{4370772,6425489}, we can evaluate the intersection point between accuracy decrease and increase in Fig. \ref{fig:drift}. If the consecutive loss performance exceeded a certain threshold, then drift warning status triggered. We measured the output performance from the new concept output and compared with the previous output. If met certain criteria, then the new AOS-ELM is committed. Otherwise, the previous AOS-ELM is rolled back. 

\subsection{MNIST and MNIST+USPS  Result}
\label{sec43}

\begin{figure}[!t]
	\centering
		\centerline{\includegraphics[width=3.25in]{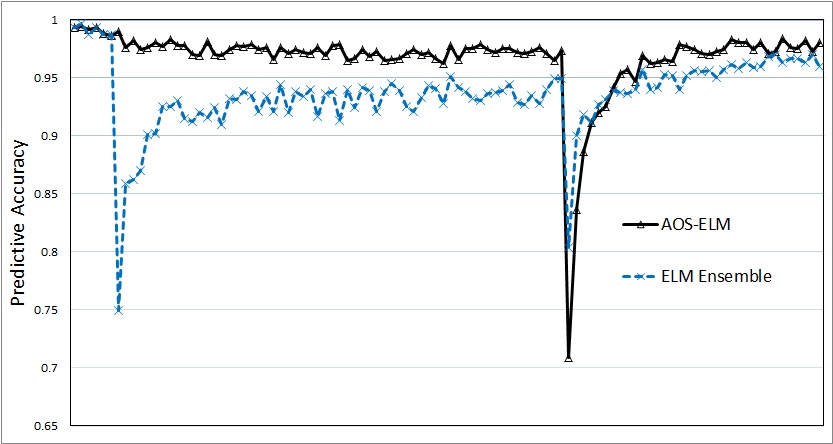}}
		\caption{Predictive accuracy of AOS-ELM (black line) over sequential data  for  MNIST+USPS compared with ELM ensemble (blue dash line)}
	\label{fig:mnist-usps1}
\end{figure}

We measured the testing accuracy based on holdout test data  by $10 \times$ experiment trials. The results are as follows:

\begin{enumerate}

\item Experiment 1 - Virtual drift.

\noindent 
The AOS-ELM of $[X_{grey}X_{HOG}]$  has Cohen's kappa of testing accuracy 95.72 (0.21) \% approximated to its non-adaptive ELM and offline ELM of $[X_{grey}X_{HOG}]$ version with the same hidden nodes number $L=2000$.  It  has better accuracy than single attribute  [$X_{grey}$] or [$X_{HOG}$] only (See Table \ref{table:442_mnist}). It proves our explanation in the theoretical background on Section \ref{sec3}. 

\textbf{Note:} We set $L_0=200$ for  [$X_{HOG}$] ELM based on the same ratio between number of input nodes with hidden nodes of [$X_{grey}$] ELM.  

\item Experiment 2 - Real drift

\noindent 
The final result as showed in Table \ref{table:443_mnist}, the AOS-ELM  has better Cohen's kappa performance for all concepts than ELM ensemble and little exceed to its non-adaptive and offline ELM. (Table \ref{table:443_mnist}). 

As well as in the split composition, the AOS-ELM with $\delta L$ increase has better performance in forgetting capability than the AOS-ELM with no $\delta L$ increase (See Table \ref{table:unbalance_mnist2}).

\item Experiment 3 - Hybrid drift

\noindent 
The final result in Table \ref{table:444_mnist}, the AOS-ELM  has better Cohen's kappa performance for HD than ELM ensemble and approximate to its non-adaptive and offline ELM.

\item Experiment 4 - MNIST+USPS Transfer Learning

The AOS-ELM  has better Cohen's kappa performance for both numeric and alphabet concepts than ELM ensemble (See Table \ref{table:mnistusps}) and approximate to its non-adaptive and offline ELM. The AOS-ELM shows better recovery time than ELM ensemble in Fig. \ref{fig:mnist-usps1}.

\end{enumerate}

\begin{table}[!t]

    \caption{Average testing accuracy and Cohen's kappa in \%  for MNIST VD experiment (Other ELM parameters are same: ROS, $\delta L=0$, $\delta N=1000$) with $10\times trials$.}
    \footnotesize
    \centering
    \subfloat[Benchmark result - non-adaptive OS-ELM and Offline ELM  \label{table:442_mnist0}]{% 
\begin{tabular}{ |p{1.25cm}|p{0.8cm}|p{1.3cm}|p{1.3cm}|p{1.75cm}| } 
\hline
Performance & ELM Method & $[X_{grey}]$ ($L=2000$)  & $[X_{HOG}]$ ($L=200$) & $[X_{grey}X_{HOG}]$ ($L=2000$)\\
\hline
\hline
\multirow{2}{4em}{Testing accuracy} & OS-ELM & $95.32\pm0.12$ & $94.64\pm0.15$ & $\mathbf{96.86}\pm0.13$ \\\cline{2-5} 
& Offline ELM & $95.33\pm0.13$ & $94.66\pm0.15$ & $\mathbf{96.85}\pm0.06$ \\\cline{1-5}  
\multirow{2}{4em}{Cohen's kappa} & OS-ELM & 94.80 (0.24) & 94.04 (0.25) & \textbf{96.51 (0.19)} \\\cline{2-5} 
& Offline ELM & 94.81 (0.23)& 94.06 (0.25) & \textbf{96.50 (0.19)} \\ \cline{1-5} 
\end{tabular}
}
\\
\subfloat[VD Experiment -  AOS-ELM ($L=2000$) \label{table:442_mnist}]{% 
\begin{tabular}{ |c|l|l| } 
\hline
Drift  &   Testing accuracy & Cohen's kappa\\
\hline
\hline
  &  &  \\
MNIST$[X_{grey}]$ $\genfrac{}{}{0pt}{}{\ggg}{VD}$ MNIST$[X_{grey}X_{HOG}]$  &  \vspace{0.1cm}$\mathbf{96.15}\pm 0.08$ &   $\mathbf{95.72 (0.21)}$ \\
  &  &  \\
\hline
\end{tabular}
}
    \end{table}

  \begin{table}[!t]

    \caption{Average testing accuracy and Cohen's kappa in \%  for MNIST RD, HD and MNIST+USPS transfer learning experiment (Other ELM parameters are same: ROS, $L=2000$, $\delta L=0$, $\delta N=1000$) with $10\times trials$.}
    % \subref{table:unbalance_mnist} split composition and \subref{table:unbalance_mnist2} split and shuffled composition}
    \footnotesize
    \centering
    \subfloat[Benchmark result - non-adaptive OS-ELM and Offline ELM\label{table:5_mnist}]{% 
\begin{tabular}{ |p{1.75cm}|c|p{1cm}|p{1cm}|p{1cm}|p{1cm}| } 
\hline
Source & Class & \multicolumn{2}{c|}{Testing Accuracy}  & \multicolumn{2}{c|}{Cohen's kappa} \\\cline{3-6}
& & OS-ELM & Offline ELM  &  OS-ELM & Offline ELM\\
\hline
\hline
\multirow{2}{4em}{MNIST $[X_{grey}]$} & (1-6) & $95.99\pm0.15$ & $96.00\pm0.14$ &95.21 (0.30) & 95.22 (0.30)  \\\cline{2-6} 
& (7-10) & $94.30\pm0.22$ & $94.32\pm0.19$ & 92.50 (0.48) & 92.53 (0.48) \\\cline{1-6}  
\multirow{2}{4em}{MNIST $[X_{grey}X_{HOG}]$} & (1-6) & $97.59\pm0.11$ & $97.49\pm0.09$ & 97.10 (0.23) & 97.00 (0.24) \\\cline{2-6} 
& (7-10) & $95.76\pm0.26$ & $95.87\pm0.12$ & 94.40 (0.42) & 94.55 (0.42) \\\cline{1-6}  
\multirow{2}{4em}{MNIST+USPS $[X_{grey}X_{HOG}]$} & (1-10) & $96.01\pm0.10$ & $96.08\pm0.08$ & 95.56 (0.02) & 95.65 (0.02)\\\cline{2-6} 
& (A-Z) & $99.94\pm0.02$ & $99.94\pm0.02$ &99.94 (0.02) & 99.93 (0.02)  \\\cline{1-6}  
\end{tabular}
}
\\
    \subfloat[RD Experiment - ELM ensemble (3 classifiers, full memory) vs. AOS-ELM .\label{table:443_mnist}]{%
     
\begin{tabular}{ |p{1.75cm}|c|p{0.9cm}|p{0.85cm}|p{0.9cm}|p{0.85cm}| } 
\hline
Source & Concept & \multicolumn{2}{c|}{Testing Accuracy}  & \multicolumn{2}{c|}{Cohen's kappa}\\\cline{3-6}
& & ELM ensemble & AOS-ELM  &  ELM ensemble & AOS-ELM\\
\hline
\hline
\multirow{2}{4em}{MNIST $[X_{grey}]$} & $\mathbf{C_1}$ (1-6) & $94.58\pm0.17$ & $\mathbf{96.09}\pm0.12$ & 93.54 (0.35) &\textbf{95.10 (0.31)}\\\cline{2-6} 
& $\mathbf{C_2}$ (7-10) & $91.60\pm0.29$ & $\mathbf{94.34}\pm0.16$ & 89.04 (0.57) &\textbf{92.56 (0.48)} \\\cline{1-6} 
\end{tabular}
}
\\
    \subfloat[HD Experiment - ELM ensemble (3 classifiers, full memory, outdated classifier pruning) vs. AOS-ELM  \label{table:444_mnist}]{%
     
\begin{tabular}{ |p{1.75cm}|c|p{0.9cm}|p{0.85cm}|p{0.9cm}|p{0.85cm}| } 
\hline
Source & Concept & \multicolumn{2}{c|}{Testing Accuracy}  & \multicolumn{2}{c|}{Cohen's kappa}\\\cline{3-6}
& & ELM ensemble & AOS-ELM  &  ELM ensemble & AOS-ELM\\
\hline
\hline

MNIST $[X_{grey}]$ & $\mathbf{C_3}$ (1-6) & $94.48\pm0.33$ & $\mathbf{97.01}\pm0.18$ & 93.42 (0.35) &\textbf{96.42 (0.26)}\\\cline{2-6} 
MNIST $[X_{grey}X_{HOG}]$ & $\mathbf{C_4}$ (7-10) & $92.29\pm0.36$ & $\mathbf{96.05}\pm0.19$ & 89.95 (0.55) &\textbf{94.78 (0.40)}\\\cline{1-6}    
\end{tabular}
    }
    \\
    \subfloat[MNIST+USPS Experiment - ELM ensemble (5 classifiers, full memory, outdated classifier pruning, ) vs. AOS-ELM  \label{table:mnistusps}]{%
    
\begin{tabular}{ |p{1.75cm}|c|p{0.9cm}|p{0.85cm}|p{0.9cm}|p{0.85cm}| } 
\hline
Source & Concept & \multicolumn{2}{c|}{Testing Accuracy}  & \multicolumn{2}{c|}{Cohen's kappa}\\\cline{3-6}
& & ELM ensemble & AOS-ELM  &  ELM ensemble & AOS-ELM\\
\hline
\hline
MNIST $[X_{grey}X_{HOG}]]$ & $\mathbf{C_5}$ (1-10) & $88.17\pm11.06$ & $\mathbf{95.91}\pm0.12$ & 86.94 (0.33) &\textbf{95.46 (0.22)}\\\cline{2-6} 
USPS $[X_{grey}X_{HOG}]$ & $\mathbf{C_6}$ (A-Z) & $99.80\pm0.05$ & $\mathbf{99.95}\pm0.03$ & 99.79 (0.40) &\textbf{99.95 (0.02)}\\\cline{1-6}    
\end{tabular}
}
    \end{table}
  
\subsection{The effect of hidden nodes increase}
\label{sec44}

% We studied the AOS-ELM with no hidden nodes increase has good performance for VD, RD and HD. However, it becomes the question what the effect of hidden nodes increase in AOS-ELM. The purpose of hidden nodes increase in ELM is to improve the generalisation performance. In a non-stationary learning, however, it is hard to determine a fixed large initial number of hidden nodes because it requires more resources, longer computation time, and sometimes not significant for any cases. When the training data more coming, we need to increase the hidden nodes in the sequential phase.

% In this section, we studied the effect of hidden nodes $\delta L$ increase in AOS-ELM. 
The initial size of hidden nodes $L_0$ selection is important to have good generalization performance. Some studies \cite{huang2006extreme,Huang201532} suggested for hidden node size to be equal at least to the rank value of training data. However, in a data stream,  it is hard to determine a fixed number of hidden nodes following that suggestion. The larger $L_0$ requires more computation resources and processing time, and probably not giving a significant result at the end. Thus, we have a requirement to increase $\delta L$ in sequential stage \cite{5178608}.

The experiment result in Table \ref{table:4421_mnist} shows the performance improved when certain hidden nodes size increase. We have different  $L_0$ conditions: 2000, the rank of initial training data (666), and rank of total training data (713) and multiple conditions of  $\delta L$ on the drift event: No increase, 500, 1000 and 2000 using ROS parameters. However, the larger $L_0$ has better influence than $\delta L$ increase. 

\setlength{\tabcolsep}{4pt}
\begin{table}[!t]
\footnotesize
\centering
\caption{Testing accuracy in Cohen's kappa (kappa error) in \% AOS-ELM MNIST for different $L_0$ and  $\delta L$ .}
\label{table:4421_mnist}
\begin{tabular}{ |p{1cm}|c|c|c|c|c| } 
\hline
Scenario & $L_0$ &  $\delta L=0$ &  $\delta L=500$ &  $\delta L=1000$ &  $\delta L=2000$  \\
\hline
\hline
\multirow{3}{4em}{VD} & 2000 & 95.92 (0.21) & 96.37 (0.20) & 96.83 (0.18) & \textbf{96.89 (0.18)} \\\cline{2-6}  
& 666  & 93.10 (0.27) & 95.18 (0.23) & \textbf{96.18 (0.20)} & 95.60 (0.22) \\\cline{2-6}  
& 713 &  93.30 (0.26) & 95.31 (0.22) & \textbf{96.28 (0.20)} & 96.09 (0.20) \\\cline{1-6} 
\multirow{3}{4em}{RD} & 2000  & 94.71 (0.24) & 94.93 (0.23) & 95.39 (0.22) & \textbf{95.42 (0.22)} \\\cline{2-6}  
& 630  & 91.3 (0.30) & 91.67 (0.29) & 93.61 (0.26) & \textbf{94.04 (0.25)} \\\cline{2-6}  
& 713 &  91.71 (0.29) & 92.70 (0.28) & 93.82 (0.25) & \textbf{94.23 (0.25)}  \\\cline{1-6}  
\end{tabular}
\end{table}

\begin{enumerate}

% \item Rank value of the pseudo inverse matrix.
\item 'Under-fitting' condition.

\noindent 'Under-fitting' is the condition when the model does not fit the data well enough that makes unconvergence. Based on an empirical experiment with $\delta L$ increase in the sequential phase on Table \ref{table:4422_mnist}, we investigated particular condition when the AOS-ELM classifier has a bad result. We realized the ELM performance is dependent upon finding general matrix inverse of $\mathbf{H}$. Based on orthogonal projection method in CEOS-ELM, we can employ the rank value of $\hat{\mathbf{P}}$ as evaluation parameter to detect 'under-fitting'.

The $\hat{\mathbf{P}}$ is approximation to matrix $(\mathbf{H}^{^\mathrm{T}}\mathbf{H})^{-1}$. 
% A full $Rank(\hat{\mathbf{P}})$ is equal to the hidden nodes number in an ideal condition. 
The full rank of $\mathbf{P}_{L \times L}$ is ideally equal with $L$. However, certain condition in the sequential training, e.g. poor training data or poor learning parameter selection may cause the diagonal squared matrix $\mathbf{P}$  has less diagonalizable \cite{Golub:1996:MC:248979}, thus not full rank anymore.

In the sequential learning, we can compare the $Rank(\hat{\mathbf{P}})$ before and after hidden nodes increase. The expected result is positive increment. If the rank value becomes lower after hidden nodes increase, then it has a higher probability of 'under-fitting' condition to occur. The $Rank(\hat{\mathbf{P}})$ is determined by the block size of training data, the number of hidden nodes increment, the $c$ scalar selection in ROS parameter, the activation function, input weight, and bias random assignment method. In this experiment, we focused on the block size, the number of hidden nodes and $c$ scalar selection. Using $Rank(\hat{\mathbf{P}})$ as evaluation parameter is more efficient because we do not need to compute $\mathbf{\beta}$. 

\setlength{\tabcolsep}{4pt}
\begin{table}[!t]
\footnotesize
\centering
\caption{Predictive accuracy performance in  AOS-ELM for MNIST using different parameters and  $Rank(\hat{\mathbf{P}})$ before and after $\delta L$ increase. Each experiment is repeated $100 \times$ trials to get the probability of predictive accuracy $\leq 50\%$.}
\label{table:4422_mnist}
\begin{tabular}{ |c|c|c|c|c|p{0.75cm}|p{0.75cm}|p{1.2cm}| } 
\hline
Scenario & $L_0$ & $\delta L$ & Batch Size & $c$ &  Before &  After & Pred. Acc. $\leq 50\%$\\
        %   &       &            &            &     & Before & After &  Acc. $\leq 50\%$ \\
\hline
\hline
\multirow{12}{4em}{RD} & 630 & 500 & 1000 & 10 & 630 & 1130 &  0\%  \\\cline{2-8}  
& 630 & 500 & 500 & 10 & 1130 & 1122 & \textbf{7}\%  \\\cline{2-8}  
& 630 & 500 & 100 & 10 & 1130 & 614 & \textbf{5}\%  \\\cline{2-8}  
& 630 & 500 & 10 & 10 & 640 & 640 & 0\%  \\\cline{2-8}  
& 630 & 100 & 1000 & 10 & 630 & 730 & 0\%  \\\cline{2-8}  
& 630 & 100 & 500 & 10 & 730 & 730 & 0\%  \\\cline{2-8}  
& 630 & 100 & 100 & 10 & 730 & 730 & 0\%  \\\cline{2-8}  
& 630 & 100 & 10 & 10 & 640 & 640 & 0\%  \\\cline{2-8} 
& 2000 & 1000 & 100 & 5 & 2000 & 1868 & \textbf{3}\%  \\\cline{2-8} 
& 2000 & 1000 & 100 & 1 & 2000 & 1947 & \textbf{16}\%  \\\cline{2-8} 
& 2000 & 1000 & 100 & 0.5 & 2000 & 1946 & \textbf{17}\%  \\\cline{2-8} 
& 2000 & 1000 & 100 & 0.05 & 2000 & 2100 & 0\%  \\\cline{1-8} 
\multirow{6}{4em}{VD} & 666 & 500 & 500 & 5 & 666 & 1166 & 0\%  \\\cline{2-8}  
& 666 & 500 & 100 & 5 & 666 & 1166 & 0\%  \\\cline{2-8}  
& 666 & 100 & 500 & 5 & 666 & 766 & 0\%  \\\cline{2-8}  
& 666 & 100 & 100 & 5 & 666 & 766 & 0\%  \\\cline{2-8}  
& 666 & 50 & 500 & 5 & 666 & 716 & 0\%  \\\cline{2-8} 
& 666 & 50 & 100 & 5 & 666 & 716 & 0\%  \\\cline{1-8} 
\end{tabular}
\end{table}

\item Sudden drift.

On  Table \ref{table:unbalance_mnist} and \ref{table:unbalance_mnist2}, 
the hidden nodes increase can improve the forgetting capability on sudden drift (it reduced the accuracy of the outdated concept). 

In  CEOS-ELM,  when $\delta L$ increase in the same time with drift, it makes    $\left[ \begin{array}{cc} \mathbf{H}_1 & \mathbf{0}  \\ \mathbf{H}_2 & \Delta \mathbf{H}_2 \end{array} \right]$ and the new concept target  $\left[ \begin{array}{c} 0 \\ \mathbf{t}_2 \end{array} \right] \in \mathbf{T}_2 $ in split composition, while previous concept $\left[ \begin{array}{c} \mathbf{t}_1 \\ 0 \end{array} \right] \in \mathbf{T}_1$. Thus, in the process of finding $\hat{\mathbf{\beta}}$ become simplified because  $\left[ \mathbf{H}_2  \Delta \mathbf{H}_2 \right]$ is partially trained by $\mathbf{t}_2$ only and not by $\mathbf{t}_1$. Thus, it reduced the generalization capability of $\left[ \mathbf{H}_2  \Delta \mathbf{H}_2 \right]$ to recognize $\mathbf{T}_1$ problem.

\begin{table}[!t]
\footnotesize
\centering
    \caption{Average Testing accuracy in \% for RD experiment. For SEA, we started from  $\mathbf{C_2}$ to $\mathbf{C_4}$. For MNIST, we started from $\mathbf{C_1}$ to $\mathbf{C_2}$.}
    
    \subfloat[Sudden drift effect  caused by  split training composition with hidden nodes $\delta L$ increase.  \label{table:unbalance_mnist}]
       {% 
      \begin{tabular}{|l|p{1.5cm}|c|c|c| } 
\hline
Data & AOS-ELM & $\delta L$ & Concept &  Testing Accuracy \\
\hline
\hline
\multirow{4}{4em}{SEA} & 
\multirow{4}{4em}{$L_0=3000$} 
& \multirow{2}{*}{0} 
&  $\mathbf{C_2}$ & $\mathbf{90.00}\pm0.59$ \\\cline{4-5} 
& & & $\mathbf{C_4}$ & $\mathbf{90.24}\pm0.61$ \\
\cline{3-5}
& & \multirow{2}{*}{500} 
&  $\mathbf{C_2}$ & $66.29\pm1.12$ \\\cline{4-5} 
& & & $\mathbf{C_4}$ & $\mathbf{90.12}\pm0.52$ \\ \cline{1-5} 
\multirow{4}{4em}{MNIST} & 
\multirow{4}{4em}{$L_0=2000$} 
& \multirow{2}{*}{0} 
& $\mathbf{C_1}$ & $\mathbf{96.42}\pm0.21$ \\\cline{4-5}  
& & &  $\mathbf{C_2}$ & $\mathbf{93.68}\pm0.23$ \\
\cline{3-5}
& & \multirow{2}{*}{500} 
& $\mathbf{C_1}$  & $17.59\pm0.98$ \\\cline{4-5} 
& & &  $\mathbf{C_2}$  & $\mathbf{97.08}\pm0.15$ \\ 
\cline{1-5}
\end{tabular}}
\\
    \subfloat[MNIST RD simulation: the effect of hidden nodes $\delta L$ increase  for split and shuffled training composition ($L_0=2000$) . \label{table:unbalance_mnist2}]{%
      \begin{tabular}{ |l|c|c|c|c| } 
\hline
Data & $\delta L$ & Composition &  $\mathbf{C_1}$ &  $\mathbf{C_2}$\\
\hline
\hline
\multirow{6}{4em}{MNIST} 
&  \multirow{2}{*}{0} 
&  Split  & $\mathbf{96.42}\pm0.21$ & $\mathbf{93.68}\pm0.23$\\\cline{3-5} 
& &  Shuffled  & $\mathbf{96.09}\pm0.12$ & $\mathbf{94.34}\pm0.16$\\ 
\cline{2-5}
&  \multirow{2}{*}{500} 
&  Split  & $17.59\pm0.98$ & $\mathbf{97.08}\pm0.15$\\\cline{3-5} 
& &  Shuffled  & $\mathbf{96.53}\pm0.12$ & $\mathbf{94.29}\pm0.25$\\ 
\cline{2-5}
& \multirow{2}{*}{1000} 
& Split  & $8.65\pm1.13$ & $\mathbf{97.64}\pm0.18$ \\\cline{3-5} 
& &  Shuffled  & $\mathbf{96.74}\pm0.14$ & $\mathbf{94.78}\pm0.10$\\
\hline
\end{tabular}
    }
  \end{table}

\end{enumerate}

\section{AOS-ELM in Regression}
\label{sec46}

We can use the similar real drift scenario with output marginalization and output amplification to solve concept drift problem in regression. 
In  this experiment,  we used AOS-ELM with single input node and single output node per concept. We defined the following concept as: 
\begin{itemize}
\item $\mathbf{C_1}$ is  \textit{sinc} function with 50000 training/5000 testing;
\item $\mathbf{C_2}$ is  \textit{sinus} function with 50000 training/5000 testing;
\item $\mathbf{C_3}$ is  \textit{gaussian} function with 50000 training/5000 testing.
\end{itemize}
The sequential experiments are following drift equations :
\begin{enumerate}
    \item Experiment 1 : $\mathbf{C_1}$ $\genfrac{}{}{0pt}{}{\ggg}{RD}$ $\mathbf{C_2}$
    \item Experiment 2 : $\mathbf{C_1}$ $\genfrac{}{}{0pt}{}{\ggg}{RD}$ $\mathbf{C_2}$ $\genfrac{}{}{0pt}{}{\ggg}{RD}$ $\mathbf{C_3}$
\end{enumerate}

We presented the result on the following figures to compare the performance of each concept at the end of each training experiment. Our objective is to show the AOS-ELM regression capability to keep the previous regression concept knowledge. We select the constant value that giving the best regression result of each concept. The AOS-ELM has $L_0=100$, $\delta L=0$, and $sigmoid$ function.  More drifts occurred will weaken the older concepts. Thus, it needs larger amplifier constant value. 

\begin{figure}[!h]

	\centering
		\centerline{\includegraphics[width=3.25in]{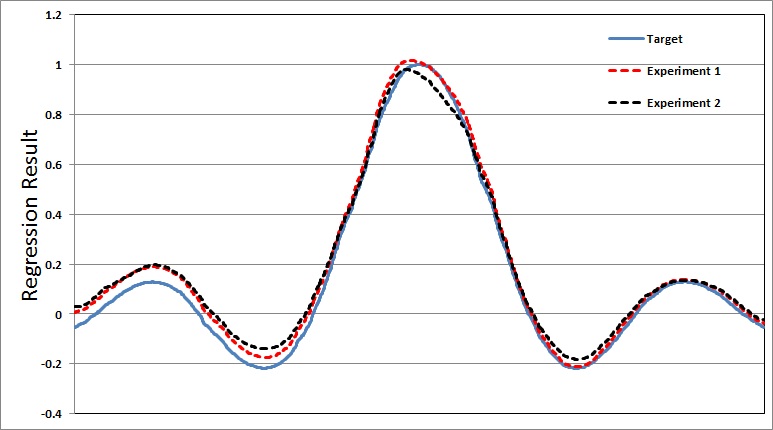}}
		\caption{Regression result of $\mathbf{C_1}$ for experiment 1 (red dash line) using constant value 4 and experiment 2  (black dash line) using constant value 8.25. }
	\label{fig:reg1}
\end{figure}	

\begin{figure}[!h]

	\centering
		\centerline{\includegraphics[width=3.25in]{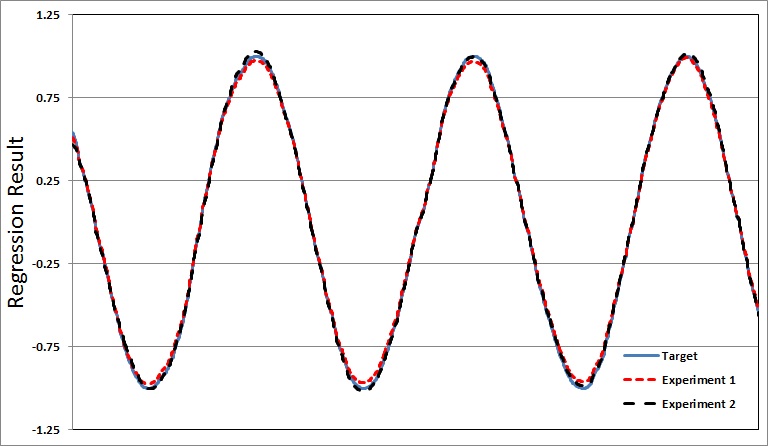}}
		\caption{Regression result of $\mathbf{C_2}$ for experiment 1 (red dash line) using constant value 1.3 and experiment 2  (black dash line) using constant value 3}
	\label{fig:reg2}
\end{figure}	

\begin{figure}[!h]

	\centering
		\centerline{\includegraphics[width=3.25in]{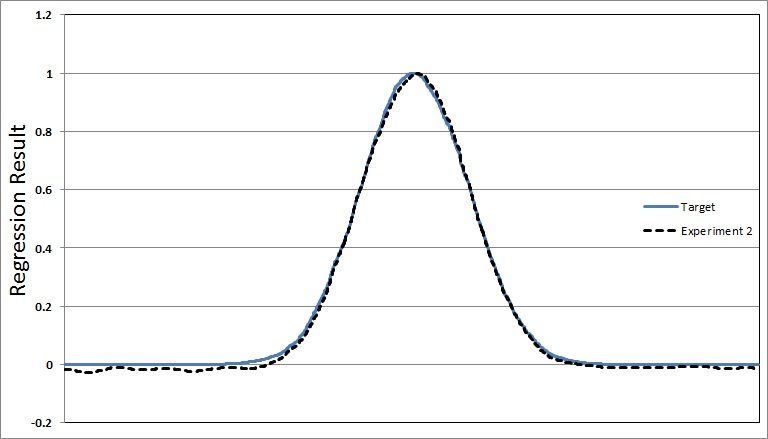}}
		\caption{Regression result of $\mathbf{C_3}$ for experiment 2  (black dash line) using constant value 1.85}
	\label{fig:reg3}
\end{figure}	

\section{Simulation in Big Data stream : Intrusion Detection System (IDS) KDD Cup 1999}
\label{aoselm-demo}

IDS is a network security technology that scans any network packet traffic to detect any potential exploits then sending the alarm or taking some active action to Intrusion Prevention System. Some machine learning methods have been applied with the hope of improving detection rates and adaptive capability \cite{sharmila-IDS}. 

\begin{figure}[!t]
	\centering
		\centerline{\includegraphics[width=3in]{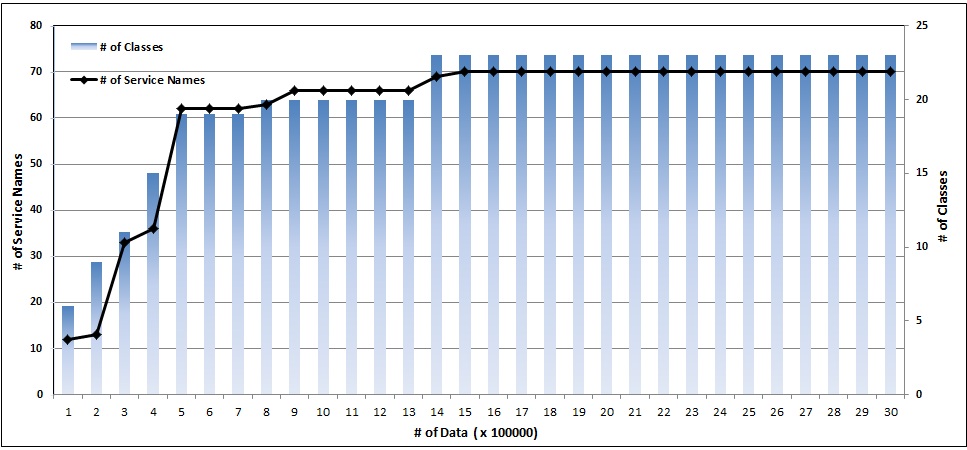}}
		\caption{The changes on Service names and number of classes following the streaming data.}
	\label{fig:ids1}
\end{figure}

In this experiment, we used  KDD Cup 1999 Competition data set. The full dataset had 4898431 network packets and grouped to be 23 classes (One  Normal class and 22 attack names based on a signature-based detection) \cite{gama-kdus}. The dataset has a control information (CI) header for delivering the data in numerical and multi-categorical values as features. We focused on service names (IP ports) attributes because they are specific differentiators for applications. The CI and the number of attack classes are not stationary. We analyzed the data set for the growing of service names and the number of class attack in the whole dataset on Fig. \ref{fig:ids1}.
The challenge in IDS dataset is imbalanced data between the classes. The highest number of data is for 'normal' class, and the lowest number is for 'spy' class (only 2 packets). To simplify the experiment, we use oversampling by adding more data based on the random normal distribution of packet signatures and under sampling approaches by dropping some samples randomly.

Based on the growing of service names and the number of classes analysis (See Fig. \ref{fig:ids1}), we designed one drift scenarios based on two concepts (Table \ref{table:ids}). 
$\mathbf{C}_{1}$ has ten classes, and 37 service names, and $\mathbf{C}_{2}$ has 23 classes and 70 service names. Total training data for each concept is 920000 packets. No data repetition from the previous event, except at the end of $\mathbf{C}_{2}$ sequential training. The composition between $\mathbf{C}_{1}$/$\mathbf{C}_{2}$ on HD event is 230000/690000.

The validation data set of $\mathbf{C}_{2}$ is selected from all packets from minority classes and randomly selected original majority classes (10422 packets). We used holdout method with $5\times$ trials. We used AOS-ELM1 for $\delta L=0$ and AOS-ELM2 for $\delta L=500$ (Other ELM parameters are same:$L_0=1000$,  NORM, \textit{sig}). The AOS-ELM result in this experiment can approximate the non-adaptive OS-ELM on $\mathbf{C}_{2}$ (See Table \ref{table:ids}). 

\begin{table}[!t]

\caption{Performance comparison  of AOS-ELM  and non-adaptive OS-ELM to  validation data set $\mathbf{C}_{2}$ in \%  . }
    \footnotesize
    \centering
    \subfloat[Benchmark result - non-adaptive OS-ELM   \label{table:ids0}]{% 
\begin{tabular}{ |c|c|p{1.5cm}|p{1.5cm}| }
\hline
Concept &  Parameters & Testing Accuracy &  Cohen's kappa \\
\hline
\hline
$\mathbf{C}_{1}$ & OS-ELM & $50.87\pm0.01$ & 45.81 (0.54) \\\cline{1-4} 
$\mathbf{C}_{2}$ & OS-ELM & $94.58\pm0.05$ & 94.03 (0.24) \\\cline{1-4} 
% $\mathbf{C}_{2}$ & OS-ELM & $72.07\pm$ & 63.97 (0.57) \\\cline{1-4} 
% $\mathbf{C}_{3}$ & OS-ELM & $97.25\pm$ & 96.97 (0.18) \\\cline{1-4} 
\hline
\end{tabular}
}
\\
\subfloat[Performance result on the drift event - AOS-ELM \label{table:ids}]{% 
\begin{tabular}{ |c|c|p{1.5cm}|p{1.5cm}| }
\hline
Drift &  Parameters & Testing Accuracy &  Cohen's kappa \\
\hline
\hline
% $\mathbf{C}_{1}$ & OS-ELM & 78.81 & 71.18 (0.54) \\\cline{1-4} 
$\mathbf{C}_{1} \genfrac{}{}{0pt}{}{\ggg}{HD} \mathbf{C}_{2}$ &  AOS-ELM1  & $92.18\pm2.73$ & 91.38 (0.29)  \\\cline{1-4}
$\mathbf{C}_{1} \genfrac{}{}{0pt}{}{\ggg}{HD} \mathbf{C}_{2}$ &  AOS-ELM2  & $94.64\pm0.06$ & 94.10 (0.24)  \\\cline{1-4}
End of full $\mathbf{C}_{2}$ &  AOS-ELM1  & $93.45\pm1.18$ & 92.78 (0.27)  \\\cline{1-4}
End of full $\mathbf{C}_{2}$ &  AOS-ELM2  & $94.57\pm0.11$ & 94.02 (0.25)  \\\cline{1-4}
% $\mathbf{C}_{1} \genfrac{}{}{0pt}{}{\ggg}{VD} \mathbf{C}_{2}$ &  AOS-ELM  & $79.10\pm$ & 71.55 (0.54)  \\\cline{1-4}
% $\mathbf{C}_{2} \genfrac{}{}{0pt}{}{\ggg}{RD} \mathbf{C}_{3}$ & AOS-ELM   & $\mathbf{98.11}\pm$ & \textbf{97.92 (0.15)}  \\\cline{1-4}
\hline
\end{tabular}
}
    \end{table}

\section{Challenges and Future Research}
\label{sec5a}

\begin{itemize}

\item We need to investigate the optimum transition space that minimize the gap to the new concept learning model. In certain case, the AOS-ELM may have the 'under-fitting' condition and require larger training data to achieve the new convergence. 

\item We need to check the consistency of AOS-ELM for different pseudo-inverse methods (E.g., Greville's method \cite{vanSchaik2015233}). 
\end{itemize}

We think some ideas for AOS-ELM  future researches: 
\begin{itemize}
    \item The need for transfer learning to solve big data problem when the distribution data changes.
    \item The  AOS-ELM integration with another ELM methods, e.g., Weighted OS-ELM for imbalanced learning \cite{journals/npl/MirzaLT13},  ELM Autoencoder (ELM-AE) \cite{Zhang20161066}, Stacked ELM \cite{6937189}, etc. 
    \item A detail systematic explanation  based on rule extraction \cite{journals/ijon/BarakatB10} for AOS-ELM in handling adaptive environment.
    % \item We need to investigate how the AOS-ELM deals with  challenges in  semi-supervised, and unsupervised \cite{6766243} problem.
\end{itemize}

\section{Conclusion}
\label{sec5}

The proposed method gives better adaptive capability than non adaptive OS-ELM and CEOS-ELM in term of retaining the recognition performance when handling concept drifts. It uses a simple line of code and easy to deploy especially for consecutive drifts, compared with adaptive ensemble methods. While most adaptive classifiers work differently for each virtual, real drift, and hybrid drift scenarios, the AOS-ELM tackles those drifts through simple block matrix reconstruction and rank evaluation. 

AOS-ELM satisfied the requirement criteria in term of accuracy,  simplicity,  fast and flexible. However, in certain VD and HD cases, the AOS-ELM accuracy may not exceed the non adaptive sequential ELM, which include the future training data. In RD cases, the AOS-ELM has better accuracy. In a real data implementation, the non-adaptive ELM is better and preferred when we know exactly the future behavior of data. However, we can not predict it precisely. We believe using larger training data, the AOS-ELM performance will approximate the expected value of non-adaptive sequential ELM or offline ELM, which use the future training data. The AOS-ELM can also add learning adaptation function to the previous offline learning model. It makes AOS-ELM an excellent choice for the unpredictable situation.

The AOS-ELM  tackles sudden drift change type as well as recurrent context change type. The output marginalization strategy is implemented by simply shifting the output nodes that belonging to the latest concept. The AOS-ELM does need to increase the hidden nodes to improve the forgetting capability for sudden drift change type. To make sure the convergence to the expected learning model, we proposed the rank value of the pseudo inverse autocorrelation hidden nodes matrix as evaluation parameter to prevent 'under-fitting' condition that makes the accuracy performance dropped. 

We can consider the AOS-ELM as another type of ELM ensemble formation using shared and interconnected hidden nodes between ensemble members. We can implement the AOS-ELM in similar fashion compared to the ELM ensemble for adaptive learning scheme, but with better performance, simpler and more resource efficient. However, the AOS-ELM does have some drawbacks. Any hidden node changes could impact all notions. 

%\nolinenumbers

%This is where your bibliography is generated. Make sure that your .bib file is actually called library.bib
\bibliography{library}

%This defines the bibliographies style. Search online for a list of available styles.
\bibliographystyle{abbrv}

\end{document}